\pdfoutput=1

\PassOptionsToPackage{dvipsnames,svgnames,table}{xcolor}
\documentclass[11pt]{article}
\usepackage{authblk}
\usepackage[]{EMNLP2023}

\usepackage{hyperref}
\usepackage{adjustbox}
\usepackage{booktabs}
\usepackage{graphicx}
\usepackage{enumitem}
\usepackage{xcolor}
\usepackage{multicol}
\usepackage{amsmath}
\usepackage{bold-extra}
\usepackage{subcaption}
\usepackage{float}
\usepackage{nicefrac}
\usepackage{color, colortbl}
\usepackage{times}
\usepackage{latexsym}
\usepackage{makecell}
\usepackage{amssymb}
\usepackage{multirow}
\usepackage{tablefootnote}
\usepackage{siunitx}
\usepackage{textcomp}
\usepackage{CJKutf8}
\usepackage{amssymb}
\usepackage{pifont}
\usepackage[dvipsnames]{xcolor}
\usepackage{array}
\usepackage{booktabs}
\usepackage{siunitx}
\usepackage{authblk}
\usepackage{url}
\renewcommand{\footnotesize}{\fontsize{8.5pt}{10.5pt}\selectfont}

\newcommand{\mytilde}{\raise.17ex\hbox{$\scriptstyle\mathtt{\sim}$}}

\usepackage[T1]{fontenc}

\usepackage[utf8]{inputenc}

\usepackage{microtype}

\usepackage{inconsolata}

\title{LLM-powered Data Augmentation for Enhanced Crosslingual Performance}

\author{
    \textbf{Chenxi Whitehouse$^{1,3,}$\thanks{$^*$Work conducted while visiting MBZUAI.}
        \quad Monojit Choudhury$^{2}$
    \quad Alham Fikri Aji$^3$} \\
    $^{1}$City, University of London \: $^{2}$Microsoft \: $^{3}$MBZUAI  \\
    \texttt{chenxi.whitehouse@city.ac.uk} \\
    \texttt{monojitc@microsoft.com} \quad \texttt{alham.fikri@mbzuai.ac.ae}
}

\begin{document}
\maketitle
\begin{abstract}

This paper explores the potential of leveraging Large Language Models (LLMs) for data augmentation in multilingual commonsense reasoning datasets where the available training data is extremely limited.
To achieve this, we utilise several LLMs, namely Dolly-v2, StableVicuna, ChatGPT, and GPT-4, to augment three datasets: XCOPA, XWinograd, and XStoryCloze.
Subsequently, we evaluate the effectiveness of fine-tuning smaller multilingual models, mBERT and XLMR, using the synthesised data.
We compare the performance of training with data generated in English and target languages, as well as translated English-generated data, revealing the overall advantages of incorporating data generated by LLMs, e.g. a notable 13.4 accuracy score improvement for the best case.
Furthermore, we conduct a human evaluation by asking native speakers to assess the naturalness and logical coherence of the generated examples across different languages.
The results of the evaluation indicate that LLMs such as ChatGPT and GPT-4 excel at producing natural and coherent text in most languages, however, they struggle to generate meaningful text in certain languages like Tamil. We also observe that ChatGPT falls short in generating plausible alternatives compared to the original dataset, whereas examples from GPT-4 exhibit competitive logical consistency. We release the generated data at \url{https://github.com/mbzuai-nlp/Gen-X}.

\end{abstract}

\section{Introduction}

The success of NLP models greatly depends on the availability and quality of training data. This poses a significant challenge for multilingual NLP, as data for languages other than English is typically limited \cite{ponti-etal-2019-modeling, joshi-etal-2020-state, whitehouse-etal-2022-entitycs}. An approach to address the data scarcity challenge is through zero-shot cross-lingual transfer or multitask training, in which a model is trained across data of diverse tasks and languages, exhibiting the capability to handle unseen tasks, particularly in larger models \cite{artetxe-schwenk-2019-massively, nooralahzadeh-etal-2020-zero, huang-etal-2021-improving-zero}. However, when aiming for task-specific objectives, a smaller, fine-tuned model dedicated to that particular task often outperforms larger general-purpose, zero-shot models. In addition, a smaller task-specific model is more practical and cost-effective for training and deployment. Nevertheless, developing a powerful task-specific model becomes challenging in the absence of training data \cite{lauscher-etal-2020-zero}. 

Conversely, recent powerful Large Language Models (LLMs) excel at handling general instructions and have shown promise in data generation tasks~\cite{wang-etal-2023-self-instruct}. 
In this work, we leverage LLMs to generate synthetic data for various multilingual commonsense reasoning tasks, XCOPA \cite{ponti-etal-2020-xcopa}, XWinograd \cite{tikhonov-ryabinin-2021-heads}, and XStoryCloze \cite{lin-etal-2022-shot}, where the training data is limited even for English (see \autoref{tab:data}). 
To augment the training data, we provide LLMs with instructions and examples from the original training data, prompting them to generate new and diverse examples.
We explore the generation of synthetic data in English using different LLMs, including open-source models like  Dolly-v2\footnote{\url{https://github.com/databrickslabs/dolly}} and StableVicuna\footnote{\url{https://github.com/Stability-AI/StableLM}}, as well as ChatGPT and GPT-4.
Although the weights and capabilities of the latter two models remain undisclosed, we explore them as they extend the capability of generating texts in languages beyond English.

\begin{table}[!t]
\centering
\sisetup{table-format=4.0} 
\addtolength{\tabcolsep}{-0.8pt}
\scalebox{0.85}{
\begin{tabular}{l|rrrrrr}
\toprule
\multirow{2}{*}{\sc\textbf{Dataset}}

& \multicolumn{2}{c}{\textbf {Train}}
& \multicolumn{2}{c}{\textbf {Validation}}
& \multicolumn{2}{c}{\textbf {Test}}
  \\
  \cmidrule(lr){2-3} \cmidrule(lr){4-5} \cmidrule(lr){6-7}

  & {\textsc{en}} & {\textsc{xx}}
& {\textsc{en}}  & {\textsc{xx}}
& {\textsc{en}}  & {\textsc{xx}}
\\
  \midrule
 {XCOPA} & 400 & 0 & 100 & 100 & 500 & 500
\\
 {XWinograd}  & 1858 & 0 & 233 & 0 & 233 & 424
\\
 {XStoryCloze} & 300 & 300 & 60 & 60 &1511  & 1511 \\

\bottomrule
\end{tabular}}
\caption{Number of examples available in XCOPA, XWinograd, and XStoryCloze. \textsc{xx} denotes the average number of non-English examples per language. Since a validation split is not specified in XStoryCloze, we take 60 random examples from the train split for validation.
XWinograd has no train/validation/test split, and we follow an 80/10/10 split for the experiments.
}
\label{tab:data}
\end{table} 
We develop task-specific models by fine-tuning multilingual pre-trained language models, namely mBERT \cite{devlin-etal-2019-bert} and XLM-R \cite{conneau-etal-2020-unsupervised}, using the generated data.
We then compare their performance against models trained on a limited set of human-created data in the target language whenever available, and otherwise through zero-shot transfer learning from manually created English training data. 
Our experiments demonstrate that training the models with {\em relatively large} synthetically generated datasets yields better performance than training with {\em limited} manually-created datasets. This finding empirically confirms the utility of synthetic data generated by LLMs for improving downstream task-specific models.

We expand the multilingual data synthesis using ChatGPT and GPT-4 on XCOPA and find that generating multilingual datasets generally surpasses the effectiveness of the zero-shot cross-lingual transfer.
We further assess the quality of the generated dataset in different languages by asking native speakers to evaluate the naturalness and logical soundness of the generated dataset compared to the human-written examples. 
The annotation results reveal that while ChatGPT and GPT-4 successfully generate natural text in most languages, they struggle with generating understandable text in certain languages such as Tamil.
Moreover, a noticeable gap is observed in terms of commonsense coherence when comparing ChatGPT-generated data to human-constructed data. On the other hand, GPT-4 significantly narrows this difference.

To summarise, our work has the following key contributions:

\begin{itemize}[nosep]
  \item  Augmenting three low-resource, multilingual commonsense reasoning datasets by leveraging and prompting four LLMs;
  \item Fine-tuning smaller models, mBERT and XLMR, using the synthesised data and showcasing the practical value of the LLM-generated data;
  \item Performing an extensive analysis of the effects of various target languages in data generation and scaling, as well as a human evaluation of the naturalness and logical coherence of the data generated in various languages;
  \item Releasing the synthesised datasets for public use and reproducibility.
\end{itemize}





\section{Related Work}

\subsection*{Multilingual and Low-Resource NLP}
Recently, there has been increased attention on expanding NLP beyond English, including the development of multilingual models~\cite{devlin-etal-2019-bert, conneau-etal-2020-unsupervised, xue-etal-2021-mt5, scao2022bloom} as well as the creation of benchmarks to address multilingual challenges~\cite{conneau2018xnli, artetxe-etal-2020-cross, adelani-etal-2021-masakhaner, winata-etal-2023-nusax}. Among the prevailing challenges faced across various languages, a common theme is the scarcity of available data. 

Consequently, when data is lacking, one approach is to employ zero-shot cross-lingual transfer. Studies conducted by \citet{winata-etal-2023-nusax} have demonstrated the effectiveness of zero-shot cross-lingual transfer for related languages. Additionally, \citet{muennighoff2022crosslingual} show that models fine-tuned only with English instruction data are capable of understanding multilingual instructions.
In this work, we are tackling a similar scenario where the availability of data is limited.



\subsection*{Multilingual Data Augmentation}
\citet{lauscher-etal-2020-zero} show that few-shot can drastically increase the cross-lingual performance of small models, proving that multilingual data augmentation is an effective strategy. 
A series of works try to predict the cross-lingual accuracy of models through measurements and modelling \cite{xia-etal-2020-predicting}, and study strategies for multilingual data augmentation, such as choosing the transfer languages \cite{lin-etal-2019-choosing}, and predicting multilingual few-shot accuracy leading for optimal data augmentation approaches \cite{tool_2022}.

Many works focus on synthetic data augmentation for code-mixing, including utilising linguistic theories \cite{lee19d_interspeech, pratapa-etal-2018-language}, machine translation models \cite{tarunesh-etal-2021-machine}, parallel corpus and Wikipedia \cite{winata-etal-2019-code, whitehouse-etal-2022-entitycs}, and employing ChatGPT \cite{dai2023chataug}. 
Our work explores data augmentation on multilingual commonsense datasets with powerful instruction-tuned LLMs.

\begin{table*}[ht]
\centering
\scalebox{0.7}{
\addtolength{\tabcolsep}{0pt}
\begin{tabular}{p{6.6cm}|p{6.3cm}|p{8.5cm}}
\toprule
\sc{XCOPA}
& \sc{XWinograd}
& \sc{XStoryCloze}                \\
\midrule

\begin{minipage}[t]{\linewidth}\includegraphics[width=0.4cm]{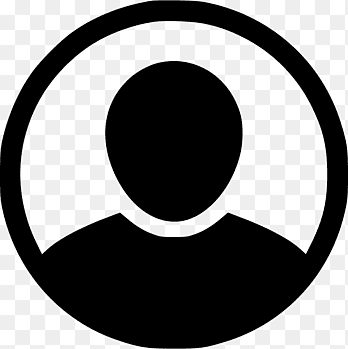} We are collecting more examples for the COPA dataset which will be used to test a system's ability of Commonsense Causal Judgments. The format of the data:\\
A premise: a statement of something that happened, and two choices that could plausibly \{\textit{occur as the result} / \textit{be the cause}\} of the premise. The correct choice is the alternative that is more plausible than the wrong choice. \\
Here are $n$ examples in \texttt{\{language\}}: \\
Example 1: \textcolor{MidnightBlue}{Premise}: The man wanted to save money. What happened as a result?  \textcolor{MidnightBlue}{Correct choice}: He cut back on making frivolous purchases. \textcolor{MidnightBlue}{Wrong choice}: He withdrew money from his savings account.  …
Example $n$: …\\
Based on the examples above, generate $m$ new examples in \texttt{\{language\}}.\end{minipage}
&
\begin{minipage}[t]{\linewidth}\includegraphics[width=0.4cm]{figures/user.png} We are collecting more examples for the Winograd Schema Challenge. Each example has a short sentence that contains two noun phrases and one pronoun replaced by ``\_'', and the challenge is to determine the referent of the pronoun, which can only be inferred from the context.\\
Here are $n$ examples of the data:  \\
Example 1:
\textcolor{MidnightBlue}{Sentence}: Harley hides from Dyna because \_ is scary. Who/What is scary? \textcolor{MidnightBlue}{Correct answer}: Dyna.
\textcolor{MidnightBlue}{Wrong answer}: Harley. … Example $n$: … \\
Based on the examples above, generate $m$ new examples. Both noun phrases in each example can be males, females, inanimate objects, or groups of people or objects. There should only be one ``\_'' in the sentence. The correct and wrong answer should be one of the noun phrases mentioned in the sentence.\end{minipage}
&
\begin{minipage}[t]{\linewidth}\includegraphics[width=0.4cm]{figures/user.png} We are collecting more examples for a story cloze dataset. Each example consists of a 4-sentence story, one correct ending sentence which is a plausible continuation of the story, and one wrong ending sentence which is logically inconsistent with the context. \\
Here are $n$ examples of the data:\\
Example 1:
\textcolor{MidnightBlue}{Sent-1}: Tina is very tired every single morning. \textcolor{MidnightBlue}{Sent-2}: She does not get enough sleep because of her two jobs. \textcolor{MidnightBlue}{Sent-3}: Tina decides to quit one of the jobs. \textcolor{MidnightBlue}{Sent-4}: She now gets enough sleep to function everyday. \textcolor{MidnightBlue}{Correct ending}: Tina is well rested. \textcolor{MidnightBlue}{Wrong ending}: Tina is more tired than ever before. … Example $n$: …\\
Based on the examples above, provide $m$ new similar examples. Requirements: 1) the story should read like a coherent story, with a specific beginning and ending, where something happens in between 2) both ending sentences should be entirely reasonable, realistic and sensible when read in isolation, and 3) both ending sentences should follow up the story by sharing at least one of the characters of the story.\end{minipage}
\\
\midrule
\begin{minipage}[t]{\linewidth}\includegraphics[width=0.4cm]{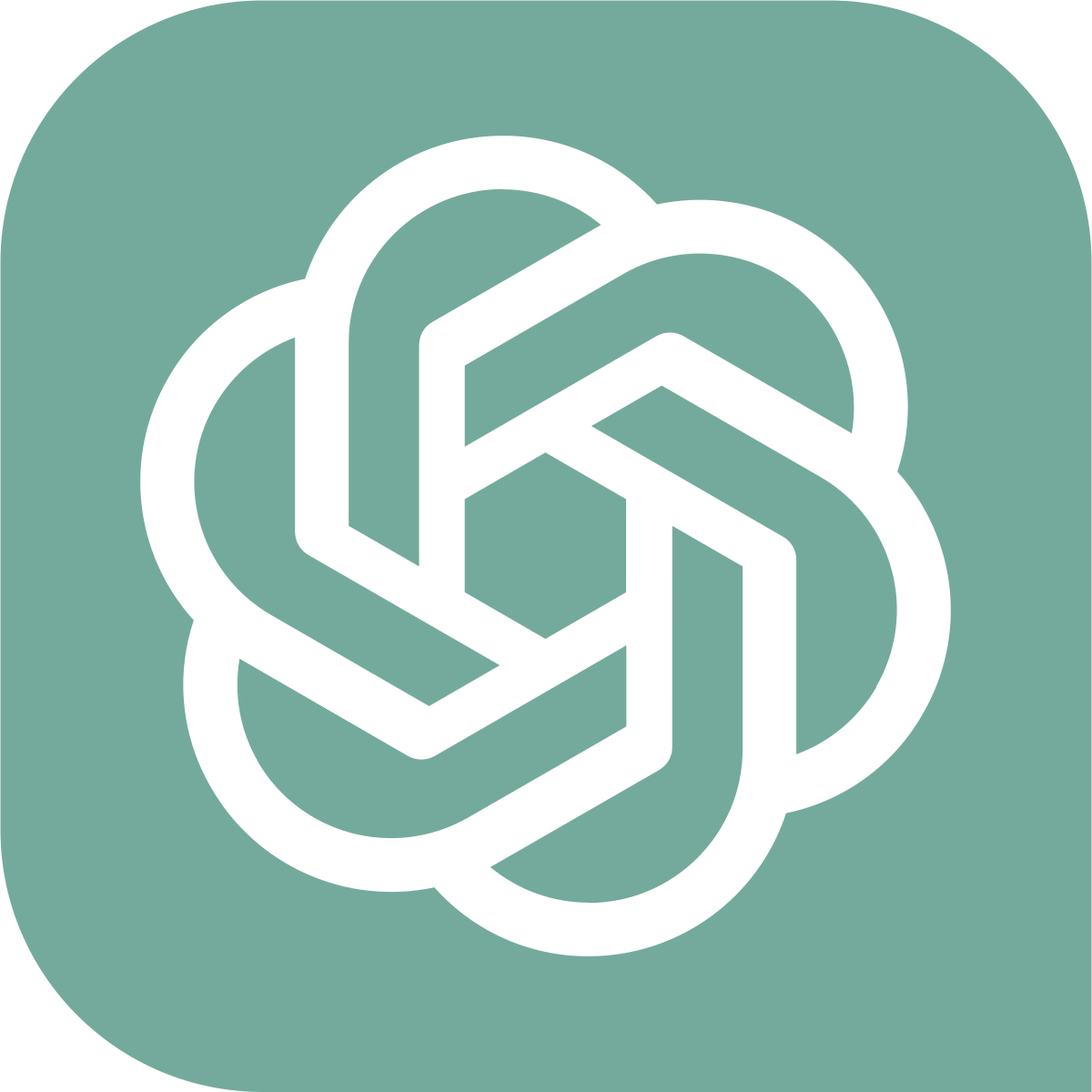}
\textcolor{MidnightBlue}{Premise}: The politician made a controversial statement. What happened as a result?
\textcolor{MidnightBlue}{Correct choice}: The politician faced criticism from the media.
\textcolor{MidnightBlue}{Wrong choice}: The politician's approval ratings increased.\\
\includegraphics[width=0.4cm]{figures/chatgpt.png}
\textcolor{MidnightBlue}{Premise}: \begin{CJK}{UTF8}{gbsn}我裤子口袋里的钥匙不见了。\end{CJK} What was the cause? \textcolor{MidnightBlue}{Correct choice}: \begin{CJK}{UTF8}{gbsn}这个口袋上有一个洞。\end{CJK} \textcolor{MidnightBlue}{Wrong choice}: \begin{CJK}{UTF8}{gbsn}裤子是新的。\end{CJK}\end{minipage}
&
\begin{minipage}[t]{\linewidth}\includegraphics[width=0.4cm]{figures/chatgpt.png} \textcolor{MidnightBlue}{Sentence}: Sam gave Andrew the book because \_ had already read it. Who/What had already read the book?
\textcolor{MidnightBlue}{Correct answer}: Sam.
\textcolor{MidnightBlue}{Wrong answer}: Andrew. \\
\includegraphics[width=0.4cm]{figures/chatgpt.png} \textcolor{MidnightBlue}{Sentence}: The dog chased the cat , but \_ was too fast. Who/What was too fast? \textcolor{MidnightBlue}{Correct answer}: the cat. \textcolor{MidnightBlue}{Wrong answer}: The dog.\end{minipage}
&
\begin{minipage}[t]{\linewidth}\includegraphics[width=0.4cm]{figures/chatgpt.png} \textcolor{MidnightBlue}{Sent-1}: Jordan was a high school student who wanted to become a doctor. \textcolor{MidnightBlue}{Sent-2}: He spent all his free time studying biology and chemistry. \textcolor{MidnightBlue}{Sent-3}: One day, his school hosted a science fair competition. \textcolor{MidnightBlue}{Sent-4}: Jordan's project won first place. \textcolor{MidnightBlue}{Correct ending}: Jordan went on to study medicine in college. \textcolor{MidnightBlue}{Wrong ending}: Jordan gave up his dream of becoming a doctor.\end{minipage}  \\
\bottomrule
\end{tabular}
}
\caption{Examples of instructions and LLM-responses (ChatGPT) for XCOPA, XWinograd, and XStoryCloze.
}
\label{tab:instructions}
\end{table*}

\section{Dataset Augmentation}


Our experiments use XCOPA, XWinograd, and XStoryCloze, which are selected due to (1) the limited availability of training data and (2) commonsense reasoning datasets present greater challenges for data synthesis.
\autoref{tab:data} summarises the statistics of the three datasets.

\paragraph{XCOPA} is a cross-lingual Choice of Plausible Alternatives dataset that translates and re-annotates the validation and test sets of English (EN) COPA \cite{roemmele2011choice} into 11 target languages (ET: Estonian, HT: Haitian Creole, ID: Indonesian, IT: Italian, QU: Quechua, SW: Swahili, TA: Tamil, TH: Thai, TR: Turkish, VI: Vietnamese, and ZH: Chinese).\footnote{\url{https://huggingface.co/datasets/xcopa}}
Each instance consists of a premise, a question
(\textit{cuase}/\textit{result}), and two alternatives. The task is to predict the more plausible alternative.

\paragraph{XWinograd} expands the original English Winograd Schema Challenge (WSC) \cite{levesque2012winograd} to five other languages (FR: French, JA: Japanese, PT: Portuguese, RU: Russian, and ZH),\footnote{\url{https://huggingface.co/datasets/Muennighoff/xwinograd}}
which consists of pronoun resolution problems aiming to evaluate the commonsense reasoning ability of a machine.
Given a statement with two noun phrases and a pronoun, the challenge of WSC is to determine the referent of the pronoun, which can only be inferred from the context.

\paragraph{XStoryCloze} is collected by
\citet{lin-etal-2022-shot},
where the validation split of the original English StoryCloze dataset \cite{mostafazadeh-etal-2016-corpus} is translated into 10 other typologically diverse languages (RU, ZH, ES: Spanish, AR: Arabic, HI: Hindi, ID, TE: Telugu, SW, EU: Basque, and MY: Burmese).
Each example consists of a four-sentence commonsense story, a correct ending, as well as a wrong ending.

\subsection{LLMs for Data Generation}
\label{sec:data_gen}
Our preliminary experiments reveal that language models that are specifically fine-tuned on downstream NLP tasks, such as BLOOMZ~\cite{scao2022bloom} and Flan-T5~\cite{chung2022scaling}, struggle to follow the complex instructions. 
Conversely, more recent LLMs such as Dolly-v2, StableVicuna,
ChatGPT, and GPT-4,
which are designed to handle more intricate and general-purpose instructions, have demonstrated success in following our instructions for data generation.
ChatGPT and GPT-4 also stand out with the capability of generating examples in non-English languages. 

 We explore synthetic data generation with the four aforementioned LLMs, balancing between open-access models and closed models (see \S\ref{sec:main-results}).
Specifically, we use \texttt{dolly-v2-12b},\footnote{Model details are included in \autoref{sec:models_urls}.} which is derived from EleutherAI’s Pythia-12b \cite{biderman2023pythia} and fine-tuned on a \mytilde15K instructions generated by Databricks employees; and \texttt{StableVicuna-13B}, an RLHF (reinforcement learning from human feedback) fine-tuned Vicuna model on various conversational and instructional datasets - Vicuna is an open-source LLaMA model \cite{touvron2023llama} fine-tuned on user-shared conversations collected from ShareGPT.\footnote{\url{https://github.com/lm-sys/FastChat}}

\subsection{Instructions and Responses}

\label{sec:data}
%
We utilise LLMs to generate synthetic examples for all datasets by prompting them.
We construct instructions using the descriptions from the dataset papers as a reference and provide LLMs with some examples, randomly sampled from the  \textit{train (+validation)} split of the original dataset, then ask LLMs to generate similar data points.
We experiment with various instructions and evaluate the synthesised data on a smaller scale, update the instructions based on the errors, and then choose the best instruction to generate the final datasets. 

The final instructions and responses are in \autoref{tab:instructions}. 
Our data generation process comprises the following key steps:
(1) We establish the desired total number of examples to generate. This quantity can be determined by various factors such as budget constraints, a fixed ratio concerning the original dataset, etc.
(2) We proceed to generate examples through the following iterative process:
(a) To ensure diversity,\footnote{An analysis of the diversity of the generation as well as topic coverage is included in \autoref{sec:diversity}.} we randomly sample a set of $n$ examples from the training datasets.
(b) We append these sampled examples to the instructions and prompt the model to generate an additional set of $m$ new examples.
(c) Afterwards, we perform post-processing and only add valid and unique examples to the generated set.
Typically, the values of $n$ and $m$ are set to 5 to 10.


\begin{table}[!t]
\centering
\scalebox{0.85}{
\addtolength{\tabcolsep}{-3pt}
\begin{tabular}{l|ccc}
    \toprule
         \textbf{Model} &  \textbf{XCOPA} &  \textbf{XWinograd} & \textbf{XStoryCloze} \\
    \midrule
       \sc  Dolly-v2 & 41.6\% & 22.4\% & 41.2\% \\
        \sc StableVicuna & 36.1\%& 33.8\% & 36.1\% \\
       \sc  ChatGPT & 86.4\% & 43.8\% & 77.6\%\\
       \sc  GPT-4 & 89.7\% & 85.0\% & 89.3\%\\
    \bottomrule
\end{tabular}}
\caption{Generation Success Rate in English (valid examples obtained / total examples requested) with different LLMs on the three datasets. }
\label{tab:success}
\end{table}

\begin{table*}[t]
\centering
\scalebox{0.8}{
\addtolength{\tabcolsep}{-2pt}
\begin{tabular}{ll|cllcllcll}
\toprule
{\multirow{2}{*}{\textbf{\parbox{0.12\textwidth}{Fine-tuned Model}}}} &
{\multirow{2}{*}{\textbf{\parbox{0.12\textwidth}{LLM for \\ Generation}}}} & \multicolumn{3}{c}{\sc \textbf{Xcopa}}                       
  & \multicolumn{3}{c}{\sc \textbf{XWinograd} }                
  & \multicolumn{3}{c}{\sc  \textbf{XStoryCloze}}                   \\
 
\cmidrule(lr){3-5}  \cmidrule(lr){6-8}  \cmidrule(lr){9-11}
& & 
 {\textit{ORI}}\textcolor{MidnightBlue}{$_{ 400}$} 
 & {\textit{GEN}}\textcolor{MidnightBlue}{$_{ 3.7k}$} 
 & {\textit{O+G}}\textcolor{MidnightBlue}{$_{ 4.1k}$} 
 &{\textit{ORI}}\textcolor{MidnightBlue}{$_{ 1.8k}$}
 & {\textit{GEN}}\textcolor{MidnightBlue}{$_{ 2k}$}
 & {\textit{O+G}}\textcolor{MidnightBlue}{$_{ 3.8k}$} 
  &{\textit{ORI}}\textcolor{MidnightBlue}{$_{ 300}$} 
 &{\textit{GEN}}\textcolor{MidnightBlue}{$_{ 1.7k}$}
 &{\textit{O+G}}\textcolor{MidnightBlue}{$_{ 2k}$}
 \\
 \midrule
\multirow{4}{*}{mBERT}      
& \sc Dolly-v2        
& 47.9 
& 53.3\textcolor{DarkGreen}{\textsubscript{~$\uparrow$5.4}} 
& 54.0\textcolor{DarkGreen}{\textsubscript{~$\uparrow$6.1}}
& 52.9
&  \textbf{59.6}\textcolor{DarkGreen}{\textsubscript{~$\uparrow$6.7}}
&  \textbf{59.3}\textcolor{DarkGreen}{\textsubscript{~$\uparrow$6.4}}
& 65.0 
& \textbf{68.7}\textcolor{DarkGreen}{\textsubscript{~$\uparrow$3.7}}  
& 68.1\textcolor{DarkGreen}{\textsubscript{~$\uparrow$3.1}} 
\\

 & \sc StableVicuna 
 & 47.9 
 & 52.9\textcolor{DarkGreen}{\textsubscript{~$\uparrow$5.0}}
 & 54.7\textcolor{DarkGreen}{\textsubscript{~$\uparrow$6.8}}
 & 52.9 
 & 53.7\textcolor{DarkGreen}{\textsubscript{~$\uparrow$0.8}}
 & 58.5\textcolor{DarkGreen}{\textsubscript{~$\uparrow$5.6}}  
 & 65.0 
 & 64.6\textcolor{Maroon}{\textsubscript{~$\downarrow$0.4}}
 & 67.3\textcolor{DarkGreen}{\textsubscript{~$\uparrow$2.3}} 
   \\
 & \sc ChatGPT      
 & 47.9 
 & 55.0\textcolor{DarkGreen}{\textsubscript{~$\uparrow$7.1}} 
 & 54.1\textcolor{DarkGreen}{\textsubscript{~$\uparrow$6.2}} 
 & 52.9 
 & 56.0\textcolor{DarkGreen}{\textsubscript{~$\uparrow$3.1}} 
 & 58.3\textcolor{DarkGreen}{\textsubscript{~$\uparrow$5.4}} 
 & 65.0 
 & 64.3\textcolor{Maroon}{\textsubscript{~$\downarrow$0.7}} 
 & 68.3\textcolor{DarkGreen}{\textsubscript{~$\uparrow$3.3}} 
 \\
& \sc GPT-4       
& 47.9 
& \textbf{56.4}\textcolor{DarkGreen}{\textsubscript{~$\uparrow$8.5}}  
& \textbf{57.2}\textcolor{DarkGreen}{\textsubscript{~$\uparrow$9.3}}  
& 52.9 
& 54.9\textcolor{DarkGreen}{\textsubscript{~$\uparrow$2.0}}  
& 57.5\textcolor{DarkGreen}{\textsubscript{~$\uparrow$4.6}} 
& 65.0 
& 68.0\textcolor{DarkGreen}{\textsubscript{~$\uparrow$3.0}}  
& \textbf{69.8}\textcolor{DarkGreen}{\textsubscript{~$\uparrow$4.8}}    
\\
\midrule
\multirow{4}{*}{XLMR-Base} 
&\sc Dolly-v2       
&54.8 
& 58.1\textcolor{DarkGreen}{\textsubscript{~$\uparrow$3.3}}  
& 58.1\textcolor{DarkGreen}{\textsubscript{~$\uparrow$3.3}}  
& 53.5 
& 56.5\textcolor{DarkGreen}{\textsubscript{~$\uparrow$3.0}}   
&  66.3\textcolor{DarkGreen}{\textsubscript{~$\uparrow$12.8}}
& 73.0 
& 75.8\textcolor{DarkGreen}{\textsubscript{~$\uparrow$2.8}}  
& 76.5\textcolor{DarkGreen}{\textsubscript{~$\uparrow$3.5}}   
\\
&\sc StableVicuna 
& 54.8 
& 57.6\textcolor{DarkGreen}{\textsubscript{~$\uparrow$2.8}} 
& 59.3\textcolor{DarkGreen}{\textsubscript{~$\uparrow$4.5}} 
& 53.5 
& 59.0\textcolor{DarkGreen}{\textsubscript{~$\uparrow$5.5}} 
& 66.0\textcolor{DarkGreen}{\textsubscript{~$\uparrow$12.5}} 
& 73.0 
& 69.6\textcolor{Maroon}{\textsubscript{~$\downarrow$3.4}} 
& 74.2\textcolor{DarkGreen}{\textsubscript{~$\uparrow$1.2}}
\\
&\sc ChatGPT     
& 54.8 
& 58.2\textcolor{DarkGreen}{\textsubscript{~$\uparrow$3.4}} 
& 59.4\textcolor{DarkGreen}{\textsubscript{~$\uparrow$4.6}} 
& 53.5 
& 62.7\textcolor{DarkGreen}{\textsubscript{~$\uparrow$9.2}} 
& 65.9\textcolor{DarkGreen}{\textsubscript{~$\uparrow$12.4}} 
& 73.0 
& 67.4\textcolor{Maroon}{\textsubscript{~$\downarrow$5.6}} 
& 74.5\textcolor{DarkGreen}{\textsubscript{~$\uparrow$1.5}}  
\\
&\sc GPT-4        
& 54.8 
& \textbf{62.7}\textcolor{DarkGreen}{\textsubscript{~$\uparrow$7.9}} 
& \textbf{63.0}\textcolor{DarkGreen}{\textsubscript{~$\uparrow$8.2}}  
& 53.5 
& \textbf{63.3}\textcolor{DarkGreen}{\textsubscript{~$\uparrow$9.8}}
& \textbf{66.9}\textcolor{DarkGreen}{\textsubscript{~$\uparrow$13.4}} 
& 73.0
& \textbf{74.6}\textcolor{DarkGreen}{\textsubscript{~$\uparrow$1.6}} 
& \textbf{79.3}\textcolor{DarkGreen}{\textsubscript{~$\uparrow$6.3}}       
\\
 \midrule
\multirow{4}{*}{XLMR-Large} 
&\sc Dolly-v2   
&   63.0 
& 58.6\textcolor{Maroon}{\textsubscript{~$\downarrow$4.4}} 
& 65.0\textcolor{DarkGreen}{\textsubscript{~$\uparrow$2.0}} 
& 80.1 
&  \textbf{76.9}\textcolor{Maroon}{\textsubscript{~$\downarrow$3.2}} 
&  83.1\textcolor{DarkGreen}{\textsubscript{~$\uparrow$3.0}} 
& 85.0
& 84.8\textcolor{Maroon}{\textsubscript{~$\downarrow$0.2}} 
& 86.4\textcolor{DarkGreen}{\textsubscript{~$\uparrow$1.4}}   
\\
&\sc StableVicuna 
&   63.0 
&64.4\textcolor{DarkGreen}{\textsubscript{~$\uparrow$1.4}}  
& 68.7\textcolor{DarkGreen}{\textsubscript{~$\uparrow$5.7}}  
& 80.1 
& 68.2\textcolor{Maroon}{\textsubscript{~$\downarrow$11.9}}  
& 82.0\textcolor{DarkGreen}{\textsubscript{~$\uparrow$1.9}}  
& 85.0 
& 74.6\textcolor{Maroon}{\textsubscript{~$\downarrow$10.4}}  
& 84.8\textcolor{Maroon}{\textsubscript{~$\downarrow$0.2}}   
\\
&\sc ChatGPT      
& 63.0 
& 64.6\textcolor{DarkGreen}{\textsubscript{~$\uparrow$1.6}} 
& 68.1\textcolor{DarkGreen}{\textsubscript{~$\uparrow$5.1}} 
& 80.1 
& 73.2\textcolor{Maroon}{\textsubscript{~$\downarrow$6.9}} 
& 83.2\textcolor{DarkGreen}{\textsubscript{~$\uparrow$3.1}}
& 85.0 
& 77.3\textcolor{Maroon}{\textsubscript{~$\downarrow$7.7}}
& 85.8\textcolor{DarkGreen}{\textsubscript{~$\uparrow$0.8}}
\\
&\sc GPT-4       
& 63.0 
& \textbf{72.1}\textcolor{DarkGreen}{\textsubscript{~$\uparrow$9.1}}
& \textbf{72.2}\textcolor{DarkGreen}{\textsubscript{~$\uparrow$9.2}} 
& 80.1 
& 76.4\textcolor{Maroon}{\textsubscript{~$\downarrow$3.7}}
& \textbf{83.5}\textcolor{DarkGreen}{\textsubscript{~$\uparrow$3.4}}
& 85.0 
& \textbf{86.0}\textcolor{DarkGreen}{\textsubscript{~$\uparrow$1.0}} 
& \textbf{88.4}\textcolor{DarkGreen}{\textsubscript{~$\uparrow$3.4}} 
\\
\bottomrule
\end{tabular}
}
\caption{Comparison of Average Accuracy across all languages for mBERT, XLMR-Base, and XLMR-Large on XCOPA, XStoryCloze, and XWinograd. Training datasets include {\textit{ORI}} (original EN data), {\textit{GEN}} (LLM-generated EN data), and {\textit{O+G}} (both), with the number of examples used for training indicated by the subscripts. The best results obtained with the same amount of training data are highlighted in bold. Green and red subscripts denote improvement and decline in performance compared to the baseline (\textit{ORI}). See per language results in \autoref{sec:additional}.
}
 \label{tab:main}
\end{table*}


We focus on a fixed-budget scenario and first generate a total of 3-4K data points for each dataset with LLMs. 
LLMs tend to generate fewer samples than requested or inconsistent output in invalid formats.
We report the success rate for different LLMs on the three datasets in \autoref{tab:success}, which indicates that GPT-4 has the most robustness.


Among the datasets, LLMs have the lowest generation success rate for XWinograd, which is more challenging.
XWinograd requires both answers to be from the generated sentence, with only one pronoun being replaced. 
In addition, we observed pronoun inconsistency in the generated XWinograd data. Despite the requirement for interchangeable pronouns in the options, models frequently fail to comply.
For example, ``The dog bit the mailman because \_ entered the yard.'' is generated by ChatGPT with the options `The dog`'' or ``the mailman'', however, ``\_'' in the sentence cannot be replaced by the same pronoun for the given two options, hence it may make the task easier and the example is considered suboptimal.
We keep those instances in the dataset and discuss further in \S\ref{sec:human}.


\section{Experimental Setups}
\label{sec:exp}

We first generate synthetic English examples for XCOPA, XWinograd, and XStoryCloze, with Dolly-v2,  StableVicuna, ChatGPT, and GPT-4.
The size of the final filtered synthesised data for the three datasets is 3.7k, 2K, and 1.7K, respectively. 
We then fine-tune mBERT, XLMR-base, and XLMR-large with the synthesised data and compare the zero-shot cross-lingual transfer performance across different languages, where we use the original validation set in target languages.

For XCOPA, we additionally experiment with generating data points directly in non-English languages, by providing examples in the target language and specifying the language desired for the generated data (see \autoref{tab:instructions}). 
However, since no examples for \textit{cause} are included in TH and TR train/validation data (they do appear in the test split), we do not generate XCOPA for the two languages. We use ChatGPT and GPT-4 for multilingual synthetic data generation, as both Dolly-v2 and StableVicuna exhibit limitations in effectively generating multilingual text.
The size of the multilingual synthesised data is \mytilde3.6K in each language.

We fine-tune models on all datasets as multiple-choice tasks\footnote{In our preliminary experiments, we find that formulating XWinograd as a binary text classification results poorly, in line with the observation from \citet{liu-etal-2020-precise} that the task formulation is essential to the performance of Winograd.} by searching best learning rate from \{$5e^{-6}$, $10e^{-6}$\}, and batch size from \{8, 16, 32\}.
All the fine-tuning experiments are conducted on a single 40G A100. 
For generating data with Dolly-v2 and StableVicuna, we use 2$\times$40G A100. 

\section{Results and Discussion}
\label{sec:results}
This section presents the main results of fine-tuned models on the three datasets and compares performance with generated data in different LLMs, languages, and scales.

\begin{table*}[t]
\centering
\scalebox{0.8}{
\addtolength{\tabcolsep}{0pt}
\begin{tabular}{lll|cccccccccc}
\toprule
\textbf{Fine-tuned} &
  \textbf{LLM} &
  \textbf{Training data} &
  \textbf{AVG} &
  \textbf{EN} &
  \textbf{ET} &
  \textbf{HT} &
  \textbf{ID} &
  \textbf{IT} &
  \textbf{SW} &
  \textbf{TA} &
  \textbf{VI} &
  \textbf{ZH} \\
  \midrule
 &
  \sc Baseline &
  \small{$ORI$} &
  \cellcolor[HTML]{FFFFFF}47.2 &
  \cellcolor[HTML]{FFFFFF}53.8 &
  \cellcolor[HTML]{FFFFFF}44.2 &
  \cellcolor[HTML]{FFFFFF}48.6 &
  \cellcolor[HTML]{FFFFFF}47.2 &
  \cellcolor[HTML]{FFFFFF}46.2 &
  \cellcolor[HTML]{FFFFFF}45.4 &
  \cellcolor[HTML]{FFFFFF}48.4 &
  \cellcolor[HTML]{FFFFFF}43.6 &
  \cellcolor[HTML]{FFFFFF}47.4 \\
  \cmidrule(l){2-13}
 &
   &
  \small{$GEN_{EN}+ORI$} &
  \cellcolor[HTML]{A2DABF}54.6 &
  \cellcolor[HTML]{AEDFC7}59.6 &
  \cellcolor[HTML]{83CDA9}56.4 &
  \cellcolor[HTML]{B9E3CF}53.6 &
  \cellcolor[HTML]{A8DCC3}53.8 &
  \cellcolor[HTML]{B7E2CD}51.4 &
  \cellcolor[HTML]{AADDC5}51.6 &
  \cellcolor[HTML]{E3F4EC}50.4 &
  \cellcolor[HTML]{88CFAD}55.0 &
  \cellcolor[HTML]{85CEAB}59.2 \\
 &
   &
    \small{$GEN_{XX}+ORI$} &
  \cellcolor[HTML]{94D4B5}56.8 &
  \cellcolor[HTML]{AEDFC7}59.6 &
  \cellcolor[HTML]{73C79E}58.8 &
  \cellcolor[HTML]{ABDDC5}54.6 &
  \cellcolor[HTML]{98D6B8}56.2 &
  \cellcolor[HTML]{70C69C}61.2 &
  \cellcolor[HTML]{96D5B7}54.6 &
  \cellcolor[HTML]{B7E2CD}53.6 &
  \cellcolor[HTML]{9CD7BA}52.0 &
  \cellcolor[HTML]{7FCBA6}60.2 \\
 &
  \multirow{-3}{*}{\sc ChatGPT} &
\small{$GEN_{EN}^{Trans}+ORI$} &
  \cellcolor[HTML]{87CFAC}58.7 &
  \cellcolor[HTML]{AEDFC7}59.6 &
  \cellcolor[HTML]{6CC499}59.8 &
  \cellcolor[HTML]{94D4B5}58.2 &
  \cellcolor[HTML]{6CC499}62.8 &
  \cellcolor[HTML]{72C69D}61.0 &
  \cellcolor[HTML]{A4DAC0}52.6 &
  \cellcolor[HTML]{9CD7BA}56.8 &
  \cellcolor[HTML]{73C79E}58.2 &
  \cellcolor[HTML]{84CEAA}59.4 \\
  \cmidrule(l){2-13}
 &
   &
  \small{$GEN_{EN}+ORI$} &
  \cellcolor[HTML]{84CDA9}59.3 &
  \cellcolor[HTML]{57BB8A}72.6 &
  \cellcolor[HTML]{73C79E}58.8 &
  \cellcolor[HTML]{C2E7D5}53.0 &
  \cellcolor[HTML]{72C69D}62.0 &
  \cellcolor[HTML]{72C69D}61.0 &
  \cellcolor[HTML]{BFE5D3}50.0 &
  \cellcolor[HTML]{B1E0C9}54.0 &
  \cellcolor[HTML]{77C8A1}57.6 &
  \cellcolor[HTML]{62C092}64.6 \\
 &
   &
    \small{$GEN_{XX}+ORI$} &
\cellcolor[HTML]{6CC499}61.8 &
  \cellcolor[HTML]{57BB8A}72.6 &
  \cellcolor[HTML]{63C093}61.2 &
  \cellcolor[HTML]{94D4B5}58.2 &
  \cellcolor[HTML]{70C69C}62.2 &
  \cellcolor[HTML]{57BB8A}66.4 &
  \cellcolor[HTML]{84CEAA}57.4 &
  \cellcolor[HTML]{C5E8D7}53.4 &
  \cellcolor[HTML]{57BB8A}63.0 &
  \cellcolor[HTML]{72C69D}61.8 \\
\multirow{-7}{*}{mBERT} &
  \multirow{-3}{*}{\sc GPT-4} &
\small{$GEN_{EN}^{Trans}+ORI$} &
  \cellcolor[HTML]{6EC49A}62.6 &
  \cellcolor[HTML]{57BB8A}72.6 &
  \cellcolor[HTML]{74C79F}58.6 &
  \cellcolor[HTML]{A8DCC3}55.2 &
  \cellcolor[HTML]{5ABD8C}65.6 &
  \cellcolor[HTML]{57BB8A}65.4 &
  \cellcolor[HTML]{9CD7BA}53.8 &
  \cellcolor[HTML]{76C8A0}62.6 &
  \cellcolor[HTML]{57BB8A}64.6 &
  \cellcolor[HTML]{5DBE8E}65.4 \\
  \cmidrule{1-13}
 &
  \sc Baseline &
  \small{$ORI$} &
  \cellcolor[HTML]{FFFFFF}55.6 &
  \cellcolor[HTML]{FFFFFF}57.6 &
  \cellcolor[HTML]{FFFFFF}54.6 &
  \cellcolor[HTML]{FFFFFF}50.6 &
  \cellcolor[HTML]{FFFFFF}59.6 &
  \cellcolor[HTML]{FFFFFF}54.8 &
  \cellcolor[HTML]{FFFFFF}55.0 &
  \cellcolor[HTML]{FFFFFF}53.4 &
  \cellcolor[HTML]{FFFFFF}54.8 &
  \cellcolor[HTML]{FFFFFF}59.6 \\
  \cmidrule(l){2-13}
 &
   &
  \small{$GEN_{EN}+ORI$} &
  \cellcolor[HTML]{C4E7D6}59.8 &
  \cellcolor[HTML]{AADDC5}63.8 &
  \cellcolor[HTML]{A5DBC1}61.6 &
  \cellcolor[HTML]{F1FAF6}51.6 &
  \cellcolor[HTML]{D5EEE2}62.6 &
  \cellcolor[HTML]{B9E3CF}59.8 &
  \cellcolor[HTML]{F4CDCD}51.6 &
  \cellcolor[HTML]{A5DBC1}60.4 &
  \cellcolor[HTML]{91D3B3}64.8 &
  \cellcolor[HTML]{DEF2E8}62.0 \\
 &
   &
    \small{$GEN_{XX}+ORI$} &
  \cellcolor[HTML]{C3E7D6}59.9 &
  \cellcolor[HTML]{AADDC5}63.8 &
  \cellcolor[HTML]{ABDDC5}60.6 &
  \cellcolor[HTML]{C2E7D5}55.0 &
  \cellcolor[HTML]{BAE3CF}64.6 &
  \cellcolor[HTML]{BCE4D1}59.6 &
  \cellcolor[HTML]{FFFFFF}54.6 &
  \cellcolor[HTML]{D5EEE2}56.4 &
  \cellcolor[HTML]{BCE4D1}59.6 &
  \cellcolor[HTML]{B7E2CD}64.8 \\
 &
  \multirow{-3}{*}{\sc ChatGPT} &
\small{$GEN_{EN}^{Trans}+ORI$} &
  \cellcolor[HTML]{B1E0CA}61.1 &
  \cellcolor[HTML]{AADDC5}63.8 &
  \cellcolor[HTML]{B4E1CB}60.0 &
  \cellcolor[HTML]{A2DABF}58.0 &
  \cellcolor[HTML]{B4E1CB}65.0 &
  \cellcolor[HTML]{ABDDC5}60.8 &
  \cellcolor[HTML]{FCF1F1}53.8 &
  \cellcolor[HTML]{A6DBC2}60.2 &
  \cellcolor[HTML]{A0D9BD}62.6 &
  \cellcolor[HTML]{A9DCC4}66.0 \\
  \cmidrule(l){2-13}
 &
   &
  \small{$GEN_{EN}+ORI$} &
  \cellcolor[HTML]{9ED8BC}63.6 &
  \cellcolor[HTML]{84CEAA}69.6 &
  \cellcolor[HTML]{96D5B7}63.8 &
  \cellcolor[HTML]{F7FCFA}51.2 &
  \cellcolor[HTML]{A1D9BE}67.2 &
  \cellcolor[HTML]{A1D9BE}62.4 &
  \cellcolor[HTML]{D0ECDF}58.4 &
  \cellcolor[HTML]{8FD2B1}63.8 &
  \cellcolor[HTML]{84CEAA}66.8 &
  \cellcolor[HTML]{93D3B4}69.4 \\
 &
   &
    \small{$GEN_{XX}+ORI$} &
  \cellcolor[HTML]{99D6B9}64.0 &
  \cellcolor[HTML]{84CEAA}69.6 &
  \cellcolor[HTML]{A1D9BE}62.2 &
  \cellcolor[HTML]{B1E0C9}56.2 &
  \cellcolor[HTML]{98D6B8}68.6 &
  \cellcolor[HTML]{98D6B8}63.8 &
  \cellcolor[HTML]{D8F0E4}57.8 &
  \cellcolor[HTML]{A2DABF}61.2 &
  \cellcolor[HTML]{84CEAA}66.8 &
  \cellcolor[HTML]{8FD2B1}70.0 \\
\multirow{-7}{*}{XLMR-Base} &
  \multirow{-3}{*}{\sc GPT-4} &
\small{$GEN_{EN}^{Trans}+ORI$} &
  \cellcolor[HTML]{9CD7BB}63.9 &
  \cellcolor[HTML]{84CEAA}69.6 &
  \cellcolor[HTML]{A5DBC1}61.6 &
  \cellcolor[HTML]{ABDDC5}56.6 &
  \cellcolor[HTML]{99D6B9}68.4 &
  \cellcolor[HTML]{8FD2B1}65.2 &
  \cellcolor[HTML]{D3EDE1}58.2 &
  \cellcolor[HTML]{A6DBC2}60.2 &
  \cellcolor[HTML]{89D0AE}66.0 &
  \cellcolor[HTML]{91D3B3}69.6 \\
  \cmidrule{1-13}
 &
  \sc Baseline &
  \small{$ORI$} &
  \cellcolor[HTML]{FFFFFF}64.4 &
  \cellcolor[HTML]{FFFFFF}71.4 &
  \cellcolor[HTML]{FFFFFF}62.8 &
  \cellcolor[HTML]{FFFFFF}51.4 &
  \cellcolor[HTML]{FFFFFF}69.0 &
  \cellcolor[HTML]{FFFFFF}65.8 &
  \cellcolor[HTML]{FFFFFF}60.6 &
  \cellcolor[HTML]{FFFFFF}62.0 &
  \cellcolor[HTML]{FFFFFF}69.4 &
  \cellcolor[HTML]{FFFFFF}66.8 \\
  \cmidrule(l){2-13}
 &
   &
  \small{$GEN_{EN}+ORI$} &
  \cellcolor[HTML]{B7E2CE}69.5 &
  \cellcolor[HTML]{B9E3CF}76.4 &
  \cellcolor[HTML]{A5DBC1}69.8 &
  \cellcolor[HTML]{F5D0D0}48.2 &
  \cellcolor[HTML]{A5DBC1}76.0 &
  \cellcolor[HTML]{A5DBC1}72.8 &
  \cellcolor[HTML]{D8F0E4}63.4 &
  \cellcolor[HTML]{AEDFC7}67.8 &
  \cellcolor[HTML]{C7E9D9}73.4 &
  \cellcolor[HTML]{8BD0AE}77.8 \\
 &
   &
    \small{$GEN_{XX}+ORI$} &
  \cellcolor[HTML]{F3FAF7}65.2 &
  \cellcolor[HTML]{B9E3CF}76.4 &
  \cellcolor[HTML]{FFFFFF}62.4 &
  \cellcolor[HTML]{CAEADB}55.2 &
  \cellcolor[HTML]{ABDDC5}75.0 &
  \cellcolor[HTML]{F4CACA}62.2 &
  \cellcolor[HTML]{F8DEDE}58.2 &
  \cellcolor[HTML]{EA9999}55.4 &
  \cellcolor[HTML]{F5D0D0}66.2 &
  \cellcolor[HTML]{95D4B6}76.2 \\
 &
  \multirow{-3}{*}{\sc ChatGPT} &
\small{$GEN_{EN}^{Trans}+ORI$} &
  \cellcolor[HTML]{DAF1E6}67.0 &
  \cellcolor[HTML]{B9E3CF}76.4 &
  \cellcolor[HTML]{F6D7D7}60.0 &
  \cellcolor[HTML]{9DD8BB}59.6 &
  \cellcolor[HTML]{F6D7D7}66.2 &
  \cellcolor[HTML]{F4FBF8}66.6 &
  \cellcolor[HTML]{FAEBEB}59.0 &
  \cellcolor[HTML]{D8F0E4}64.8 &
  \cellcolor[HTML]{B4E1CB}74.8 &
  \cellcolor[HTML]{99D6B9}75.6 \\
  \cmidrule(l){2-13}
 &
   &
  \small{$GEN_{EN}+ORI$} &
  \cellcolor[HTML]{95D5B6}73.7 &
  \cellcolor[HTML]{7CCAA4}84.6 &
  \cellcolor[HTML]{A1D9BE}70.4 &
  \cellcolor[HTML]{FBEEEE}50.0 &
  \cellcolor[HTML]{85CEAB}80.8 &
  \cellcolor[HTML]{74C79F}80.2 &
  \cellcolor[HTML]{B7E2CD}65.8 &
  \cellcolor[HTML]{8CD1AF}72.8 &
  \cellcolor[HTML]{98D6B8}78.4 &
  \cellcolor[HTML]{7AC9A2}80.4 \\
 &
   &
    \small{$GEN_{XX}+ORI$} &
  \cellcolor[HTML]{8FD2B1}74.6 &
  \cellcolor[HTML]{7CCAA4}84.6 &
  \cellcolor[HTML]{76C8A0}77.0 &
  \cellcolor[HTML]{BFE5D3}56.0 &
  \cellcolor[HTML]{7CCAA4}82.2 &
  \cellcolor[HTML]{89D0AE}77.0 &
  \cellcolor[HTML]{C2E7D5}65.0 &
  \cellcolor[HTML]{8BD0AE}73.8&
  \cellcolor[HTML]{A6DBC2}76.2 &
  \cellcolor[HTML]{7CCAA4}80.0 \\
\multirow{-7}{*}{XLMR-Large} &
  \multirow{-3}{*}{\sc GPT-4} &
\small{$GEN_{EN}^{Trans}+ORI$} &
  \cellcolor[HTML]{93D4B4}74.1 &
  \cellcolor[HTML]{7CCAA4}84.6 &
  \cellcolor[HTML]{88CFAD}74.2 &
  \cellcolor[HTML]{AEDFC7}57.2 &
  \cellcolor[HTML]{7ECBA5}82.0 &
  \cellcolor[HTML]{87CFAC}77.4 &
  \cellcolor[HTML]{E9F6F0}62.2 &
  \cellcolor[HTML]{7ECBA5}75.0 &
  \cellcolor[HTML]{B9E3CF}74.4 &
  \cellcolor[HTML]{7FCBA6}79.6\\
  \bottomrule
\end{tabular}
}
\caption{Accuracy on XCOPA.
{\footnotesize{$ORI$}} corresponds to the original data, 
{\footnotesize{$GEN_{EN}$}} and {\footnotesize{$GEN_{XX}$}} represents data generated in English and target languages.
{\footnotesize{$Trans$}} denotes translations of the English-generated data.
We show languages that are available in all settings. 
Improvement and decline in performance are represented with green and red shadows.
}
 \label{tab:xcopa}
\end{table*}

\subsection{General Result}

\label{sec:main-results}
\autoref{tab:main} presents the average accuracy of fine-tuned mBERT, XLMR-Base, and XLMR-Large models across all languages on the three datasets.
The models are trained using original data (\textit{ORI}), different LLM-generated data (\textit{GEN}), as well as a combination of both sources (\textit{O+G}) in English. 

Across different datasets, LLMs, and fine-tuned models, consistent improvements are observed when using both original and LLM-generated data. Among the models, Dolly-v2 performs the best on Xingorad when fine-tuned on mBERT, while GPT-4 achieves the highest accuracy in other settings.
The most significant improvement is shown in XWinograd with XLMR-Base, where the addition of an extra 2k datapoints leads to an average accuracy enhancement of 12.8 compared to the baseline, across all four LLMs.

When using only LLM-generated data, smaller models like mBERT and XLMR-Base generally outperform the baseline. However, with XLMR-Large, which achieves stronger baselines. e.g. >80 in XWinograd and XStoryCloze, the accuracy remains similar or even worse compared to using the original data.
GPT-4-generated data demonstrates the best robustness but still experiences a decline in performance in XWinograd when the generated data size is similar to the original data. This highlights the challenges of generating data at a human-level quality.


\subsection{Multilingual Data Generation}

We investigate whether the synthetically generated multilingual dataset outperforms training solely in English. 
We choose the XCOPA dataset and explore two settings: synthetic multilingual data by asking LLMs to generate responses in the target languages directly and translating the English-generated data to target languages with Google Translate API. 
We exclude Dolly-v2 and StableVicuna due to their limited effectiveness in generating non-English text. 
Although GPT-4 exhibits the most promising performance, it is significantly costlier compared to ChatGPT. Therefore, we also consider using ChatGPT as a contrasting experiment under resource-constrained conditions.

\autoref{tab:xcopa} shows the results for the languages that are available for all settings, excluding TR and TH (unavailable for LLM-generation, refer to \S\ref{sec:exp}), and QU (not supported by the Google Translate API).
We can see the impact of the generated data varies across different fine-tuned models and languages, aligning with the findings of \citet{kumar-etal-2022-diversity}.
Training on GPT-4 synthesised data displays consistent improvement across all scenarios and languages, except the zero-shot cross-lingual result on HT with XLMR-Large.

More fluctuating results can be observed with ChatGPT-generated data. 
A comparison between {\small{$GEN_{EN} + ORI$}} and {\small{$GEN_{XX} + ORI$}} indicates that utilising data generated in target languages generally leads to improved performance with GPT-4 generated data, as well as in base models with ChatGPT-generated data.
However, for XLMR-Large, employing ChatGPT-generated data in target languages mostly yields negative outcomes.
In languages such as TA and VI, training on generated data in the target languages results in more performance degradation compared to zero-shot cross-lingual transfer. This suggests that ChatGPT performs worse in those languages than XLMR-Large \cite{ahuja2023mega}.

Translating the English dataset generally shows overall better results than training on the data generated directly in the target languages, with the exception of XLMR-Large with GPT-4.
For SW, XLMR models fined-tuned with ChatGPT-generated data exhibit performance decline in most cases, even when the English-generated data benefits all other languages. This observation suggests that XLMR struggles with SW.
In \S\ref{sec:human} we select TA, SW, and the two best languages, ID and ZH, along with EN, for human evaluation.

Additionally, we conduct experiments adding Target Languages in Validation (TLV).
This only results in minor variations in the performance, consistent with the findings of \citet{ponti-etal-2020-xcopa}.
We include the full results in \autoref{tab:xcopa_full} in \autoref{sec:additional}.

\subsection{Dataset Scaling Up}

\begin{table}[!t]
\centering
\scalebox{0.85}{
\addtolength{\tabcolsep}{-1pt}
\begin{tabular}{l|cccc}
\toprule
\multirow{2}{*}{\textbf{Model}}

& \multicolumn{2}{c} {\small{$GEN_{EN} + ORI_{EN}$} }
& \multicolumn{2}{c}{\small{$GEN_{EN}^{Trans} + ORI_{EN}$} }
  \\
  \cmidrule(lr){2-3} \cmidrule(lr){4-5}

  & \textsc{3.7k} & \textsc{28.6k}
& \textsc{3.7k}  & \textsc{28.6k}
\\
\midrule

mBERT &
 54.3 & 56.0 & 58.0 & \textbf{60.1}
\\
XLMR-Base
& 60.1 & \textbf{61.8}
& 61.2 & 61.7

\\
XLMR-Large
 & 69.7 &\textbf{72.4} & 67.2 & 71.4 \\

\bottomrule
\end{tabular}}
\caption{Accuracy on XCOPA when scaling up the generated data to over 28K with ChatGPT. We report average results on all XCOPA languages excl. QU, since it is not available with the Google Translate API.}
\label{tab:scale}
\end{table}

We now investigate the impact of training on a larger scale of generated data on model performance. 
We focus on the XCOPA dataset and expand the generated data with ChatGPT (more budget-efficient) to 28.6k examples in English. We also compare the results of zero-shot cross-lingual transfer with translating the English-generated data to target languages.

The results in \autoref{tab:scale} demonstrate the positive impact of scaling up the generated data on model performance. Particularly, XLMR-Large exhibits the most significant improvement.


Furthermore, we conduct experiments on generating data with a fixed ratio of the original datasets and the results are included in \autoref{sec:ratio}.

\begin{figure*}[t!]
\centering
    \includegraphics[width=0.95\linewidth]{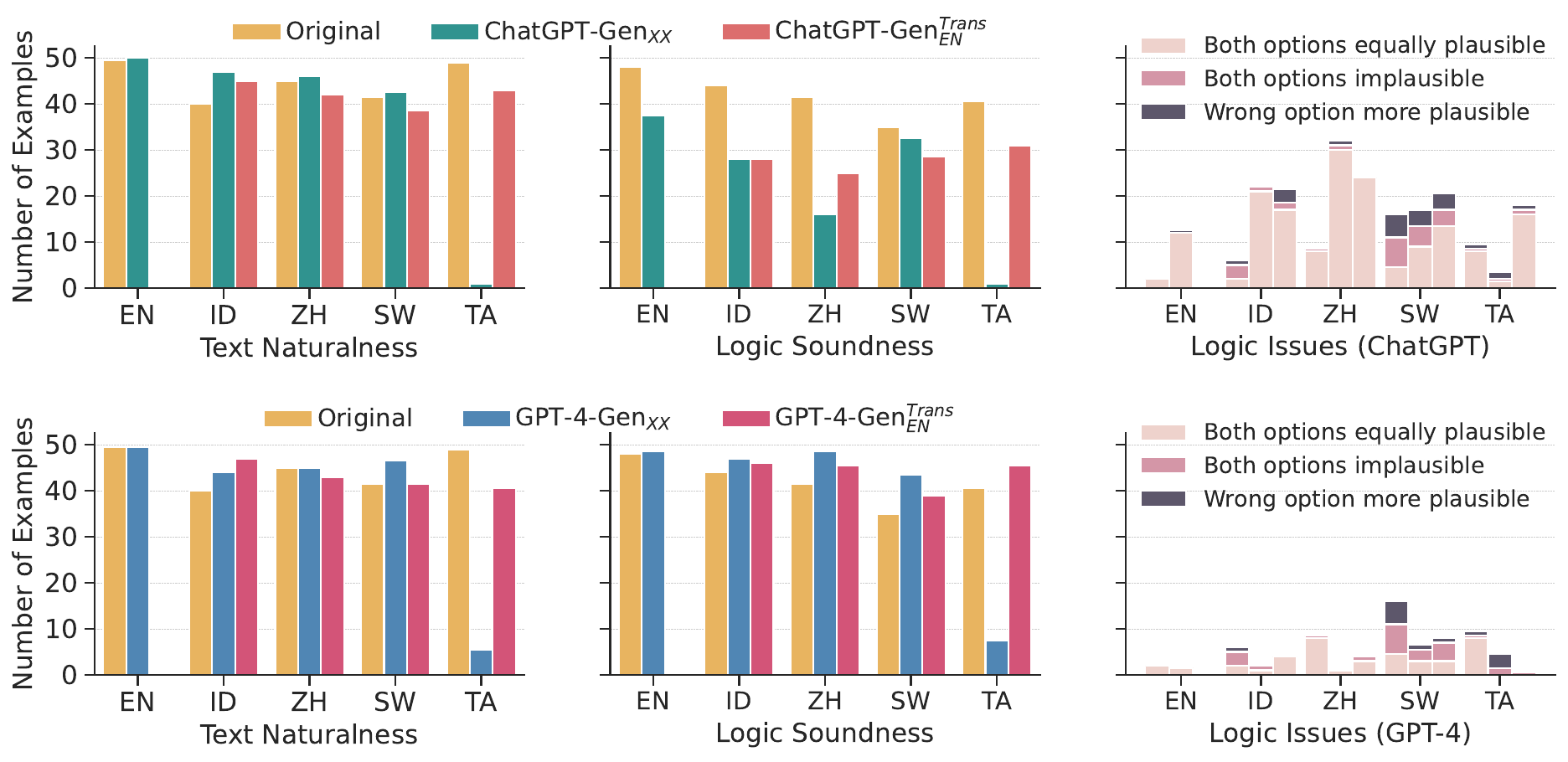}
\caption{Human evaluation of 50 random examples from the original XCOPA, ChatGPT (top) and GPT-4 (bottom) generated data in target languages, and translation of English generated data. Examples are annotated by two native speakers in each language. 
    The subplots in the last column show the logic issues of the XCOPA data, where the three bars for each language represent \textit{Oringal}, {\footnotesize{$Gen_{XX}$}}, and {\footnotesize{$Gen_{EN}^{Trans}$}} (from left to right).
}
\label{fig:human}
\end{figure*}

\section{Human Evaluation}



To better evaluate the quality of the generated datasets and compare them with the human-created data, we ask native speakers to annotate the multilingual data generated by ChatGPT and GPT-4.

For each dataset, we first select 50 generated examples in English, and then request two annotators to evaluate the examples in two categories: 
(1) \textbf{Text Naturalness}. The annotators are asked to choose one of the following options for each example:  ``the text sounds natural'', ``the text sounds awkward but understandable'', or ``the text is not understandable'', and 
(2) \textbf{Logic Soundness}. This category focuses on the commonsense aspect of the examples. The annotators are required to select the most appropriate description from: ``the correct option is (clearly) more plausible'', ``both options are equally plausible'', ``both options are implausible'', or ``the wrong option is actually more plausible''. We only ask the annotators to evaluate the logic if the text is at least understandable.

For XWinograd, we introduce an additional evaluation criterion. Annotators are asked to determine whether the two noun phrases in the examples can be replaced by the same pronoun (refer to \S\ref{sec:data}).
For XCOPA, we extend the annotations to non-English languages, where we choose the two languages that demonstrate the most notable improvement, namely ZH and ID, as well as the two languages that exhibit the least improvement or regression in performance with ChatGPT-generated data, namely TA and SW (see \autoref{tab:xcopa}).
In addition to the original examples and the generated examples in the target languages, we include 50 examples that are translated from the same English-generated examples (that were selected for annotation).

To ensure impartiality, all the examples are shuffled, and the annotators are not provided with information regarding the source of the examples (human-created, LLM-generated, or translated).

\subsection{Text Naturalness}
\label{sec:human}
\autoref{fig:human} presents the annotation results for XCOPA, averaged from two annotators for each language.
For Text Naturalness,
we can see that in EN, ID, ZH, and SW, both ChatGPT and GPT-4 achieved higher naturalness than the original dataset.
This is particularly prominent in ID, revealing the fluency issue in the original ID data in XCOPA, which is also confirmed by a native speaker.

\subsection*{Issue with Tamil}

In contrast, the performance of the TA dataset is surprisingly low, with a majority of examples classified as "not understandable." Upon consulting language experts, we have identified several main issues in Tamil, including 
(1) the insertion of redundant words with the same meaning, such as ``I will retry to try it again'' 
(2) verb agreement errors, and 
(3) the presence of uncommon and out-of-context words.

It is worth noting that generating Tamil using GPT-4 is both slow and costly. We suspect that the tokenizer for Tamil, as well as similar languages like Telugu and Kannada, are poorly trained, resulting in unusable generation in those languages.
While the low quality of the generated data could explain the significant decline in the performance of the XLMR-Large model when trained on ChatGPT-generated data in Tamil, intriguingly, models trained on Tamil data generated by GPT-4 show improvement over the baselines.

To further investigate this issue, we conduct an experiment where we fine-tune the models using only five examples from the TA examples generated by GPT-4 that are identified as natural and sound by the annotators. 
The improvement on mBERT under this setting is 50\% of the total improvement seen with the entire 3.6K TA examples. For XLMR-base and XLMR-large, 15\% and 3\% of the total improvement can be observed, respectively.
Considering that the estimated number of correct samples in the 3.6k dataset is around 360, it is plausible that training solely on those examples could raise the accuracy level, or even surpass, what we observe for the entire dataset.\footnote{We could not conduct this experiment as the entire dataset was not manually labelled.}
An intriguing question that remains to be investigated in future research is why the remaining 3.2k incorrect or unnatural examples do not negatively impact the model's performance.

The translated text is typically less natural than the original and generated data (apart from ID due to issues in the original data). This result affirms that LLMs generally excel in generating fluent text for the languages it supports.

\subsection{Logic Soundness}
In terms of logic soundness, ChatGPT falls short compared to the original dataset.
We further illustrate the categorised issues in the last column of the plots in \autoref{fig:human}.
We can see that for ChatGPT, the majority of the examples are labelled as ``both options are equally plausible'', only SW has more problematic examples with ``the wrong option is actually more plausible''. 
We suspect that this issue arises from the instruction provided (taken from the description of the original COPA dataset), which states that ``both options could be plausible, but one is more plausible.''
In some cases, ChatGPT generates two choices that are excessively similar in terms of plausibility. 
On the other hand, GPT-4 tends to generate options with more clear-cut differences in plausibility, mirroring the original data.
We note that despite the description/instruction that both alternatives could happen, both the original dataset and the data synthesised by GPT-4 tend to present one plausible and one \textit{implausible} option.

For English XWinograd and XstoryCloze, the majority of the examples in both original and generated examples are evaluated as natural and logically sound. 
For XWinograd, although more than 47 examples are evaluated to exhibit high text quality and follow commonsense logic, only 23 ChatGPT-generated examples fulfil the requirement that both noun phrases should be interchangeable with the same pronoun.
GPT-4 examples demonstrate better consistency, with 36 following this rule, whereas all original examples are found satisfactory.

\section{Conclusions}
This paper explores the effectiveness of utilising LLMs for data augmentation in cross-lingual datasets with limited training data. 
We specifically focus on commonsense reasoning tasks that are challenging for data synthesis.
Our experiments including four LLMs for data generation on three datasets, showcase enhanced cross-lingual zero-shot transfer on smaller fine-tuned task-specific language models. However, the impact varies across different datasets and languages. Notably, larger models such as XLMR-Large, which have higher baselines, demonstrate more difficulty in achieving performance improvements with LLM-generated data. Among the four LLMs, GPT-4-generated data exhibits mostly consistent superior performance.

Expanding data generation directly in target languages also shows general improvements compared to cross-lingual zero-shot with the English-generated data.
Human evaluation of the synthesised multilingual dataset shows that the ChatGPT and GPT-4 generated data demonstrate high naturalness in most languages, even surpassing the original data. However, in certain languages like TA, both models fail to generate natural text. Additionally, when assessing the logical soundness of the dataset, examples synthesised by ChatGPT reveal notable inconsistencies regarding more plausible options compared to the original human-created data. In contrast, GPT-4 exhibits a performance on par with human-written data.

In conclusion, leveraging LLMs for data augmentation shows promise. However, the choice of LLM used for data generation significantly influences the quality of the resulting data, as well as its applicability to the language under consideration. In circumstances where a more advanced model such as GPT-4 cannot be accessed, other models can be utilised, though this might result in performance difficulties in certain non-English languages - a challenge that also exists for GPT-4 - and concerns regarding logical coherence. A compelling direction for future research could involve exploring the efficacy of more recent instruction-tuned or aligned open-source LLMs, such as LLaMA 2 \cite{touvron2023llama2} or TÜLU \cite{wu2023fine}, in enhancing data augmentation.


\section*{Limitations}

We have identified the following limitations in this work:
(1)While LLMs, especially GPT-4, exhibit promising results in the context of multilingual commonsense data augmentation, they may encounter challenges when applied to extremely low-resource languages.
(2)
In order to achieve optimal performance, few-shot examples in the target language are still necessary for generating new examples. However, acquiring such examples may not always be feasible for all languages of interest.
(3)
The usage of closed models like GPT-4 is limited by licensing restrictions, and the results obtained from these models may not be reproducible. Nonetheless, the experiments conducted in this study demonstrate the potential benefits of leveraging LLMs for multilingual dataset augmentation.

\section*{Ethical Consideration}
Synthetic data generation with LLMs, especially multilingual data, should be approached with sensitivity and respect, as it reflects the linguistic, social, and cultural identity of a multilingual community. 
Since LLMs are trained on web data, they may encode biases perpetuating stereotypes, discrimination, or marginalisation of specific languages or communities. Therefore, collaboration with linguists, language experts, and community representatives is necessary to avoid the unintentional perpetuation of stereotypes and cultural insensitivity.


\bibliography{custom}

\begin{thebibliography}{39}
\expandafter\ifx\csname natexlab\endcsname\relax\def\natexlab#1{#1}\fi

\bibitem[{Adelani et~al.(2021)Adelani, Abbott, Neubig, D’souza, Kreutzer, Lignos, Palen-Michel, Buzaaba, Rijhwani, Ruder et~al.}]{adelani-etal-2021-masakhaner}
David~Ifeoluwa Adelani, Jade Abbott, Graham Neubig, Daniel D’souza, Julia Kreutzer, Constantine Lignos, Chester Palen-Michel, Happy Buzaaba, Shruti Rijhwani, Sebastian Ruder, et~al. 2021.
\newblock \href {https://doi.org/10.1162/tacl_a_00416} {Masakhaner: Named entity recognition for african languages}.
\newblock \emph{Transactions of the Association for Computational Linguistics}, 9:1116--1131.

\bibitem[{Ahuja et~al.(2023)Ahuja, Hada, Ochieng, Jain, Diddee, Maina, Ganu, Segal, Axmed, Bali et~al.}]{ahuja2023mega}
Kabir Ahuja, Rishav Hada, Millicent Ochieng, Prachi Jain, Harshita Diddee, Samuel Maina, Tanuja Ganu, Sameer Segal, Maxamed Axmed, Kalika Bali, et~al. 2023.
\newblock \href {https://arxiv.org/abs/2303.12528} {Mega: Multilingual evaluation of generative ai}.
\newblock \emph{arXiv preprint arXiv:2303.12528}.

\bibitem[{Artetxe et~al.(2020)Artetxe, Ruder, and Yogatama}]{artetxe-etal-2020-cross}
Mikel Artetxe, Sebastian Ruder, and Dani Yogatama. 2020.
\newblock \href {https://doi.org/10.18653/v1/2020.acl-main.421} {On the cross-lingual transferability of monolingual representations}.
\newblock In \emph{Proceedings of the 58th Annual Meeting of the Association for Computational Linguistics}, pages 4623--4637, Online. Association for Computational Linguistics.

\bibitem[{Artetxe and Schwenk(2019)}]{artetxe-schwenk-2019-massively}
Mikel Artetxe and Holger Schwenk. 2019.
\newblock \href {https://doi.org/10.1162/tacl_a_00288} {Massively multilingual sentence embeddings for zero-shot cross-lingual transfer and beyond}.
\newblock \emph{Transactions of the Association for Computational Linguistics}, 7:597--610.

\bibitem[{Biderman et~al.(2023)Biderman, Schoelkopf, Anthony, Bradley, O'Brien, Hallahan, Khan, Purohit, Prashanth, Raff, Skowron, Sutawika, and Van Der~Wal}]{biderman2023pythia}
Stella Biderman, Hailey Schoelkopf, Quentin~Gregory Anthony, Herbie Bradley, Kyle O'Brien, Eric Hallahan, Mohammad~Aflah Khan, Shivanshu Purohit, Usvsn~Sai Prashanth, Edward Raff, Aviya Skowron, Lintang Sutawika, and Oskar Van Der~Wal. 2023.
\newblock \href {https://proceedings.mlr.press/v202/biderman23a.html} {Pythia: A suite for analyzing large language models across training and scaling}.
\newblock In \emph{Proceedings of the 40th International Conference on Machine Learning}, volume 202 of \emph{Proceedings of Machine Learning Research}, pages 2397--2430. PMLR.

\bibitem[{Chung et~al.(2022)Chung, Hou, Longpre, Zoph, Tay, Fedus, Li, Wang, Dehghani, Brahma et~al.}]{chung2022scaling}
Hyung~Won Chung, Le~Hou, Shayne Longpre, Barret Zoph, Yi~Tay, William Fedus, Eric Li, Xuezhi Wang, Mostafa Dehghani, Siddhartha Brahma, et~al. 2022.
\newblock \href {https://arxiv.org/abs/2210.11416} {Scaling instruction-finetuned language models}.
\newblock \emph{arXiv preprint arXiv:2210.11416}.

\bibitem[{Conneau et~al.(2020)Conneau, Khandelwal, Goyal, Chaudhary, Wenzek, Guzm{\'a}n, Grave, Ott, Zettlemoyer, and Stoyanov}]{conneau-etal-2020-unsupervised}
Alexis Conneau, Kartikay Khandelwal, Naman Goyal, Vishrav Chaudhary, Guillaume Wenzek, Francisco Guzm{\'a}n, Edouard Grave, Myle Ott, Luke Zettlemoyer, and Veselin Stoyanov. 2020.
\newblock \href {https://doi.org/10.18653/v1/2020.acl-main.747} {Unsupervised cross-lingual representation learning at scale}.
\newblock In \emph{Proceedings of the 58th Annual Meeting of the Association for Computational Linguistics}, pages 8440--8451, Online. Association for Computational Linguistics.

\bibitem[{Conneau et~al.(2018)Conneau, Rinott, Lample, Williams, Bowman, Schwenk, and Stoyanov}]{conneau2018xnli}
Alexis Conneau, Ruty Rinott, Guillaume Lample, Adina Williams, Samuel Bowman, Holger Schwenk, and Veselin Stoyanov. 2018.
\newblock \href {https://doi.org/10.18653/v1/D18-1269} {{XNLI}: Evaluating cross-lingual sentence representations}.
\newblock In \emph{Proceedings of the 2018 Conference on Empirical Methods in Natural Language Processing}, pages 2475--2485, Brussels, Belgium. Association for Computational Linguistics.

\bibitem[{Dai et~al.(2023)Dai, Liu, Liao, Huang, Wu, Zhao, Liu, Liu, Li, Zhu et~al.}]{dai2023chataug}
Haixing Dai, Zhengliang Liu, Wenxiong Liao, Xiaoke Huang, Zihao Wu, Lin Zhao, Wei Liu, Ninghao Liu, Sheng Li, Dajiang Zhu, et~al. 2023.
\newblock \href {https://arxiv.org/abs/2302.13007} {Chataug: Leveraging chatgpt for text data augmentation}.
\newblock \emph{arXiv preprint arXiv:2302.13007}.

\bibitem[{Devlin et~al.(2019)Devlin, Chang, Lee, and Toutanova}]{devlin-etal-2019-bert}
Jacob Devlin, Ming-Wei Chang, Kenton Lee, and Kristina Toutanova. 2019.
\newblock \href {https://doi.org/10.18653/v1/N19-1423} {{BERT}: Pre-training of deep bidirectional transformers for language understanding}.
\newblock In \emph{Proceedings of the 2019 Conference of the North {A}merican Chapter of the Association for Computational Linguistics: Human Language Technologies, Volume 1 (Long and Short Papers)}, pages 4171--4186, Minneapolis, Minnesota. Association for Computational Linguistics.

\bibitem[{Huang et~al.(2021)Huang, Ahmad, Peng, and Chang}]{huang-etal-2021-improving-zero}
Kuan-Hao Huang, Wasi Ahmad, Nanyun Peng, and Kai-Wei Chang. 2021.
\newblock \href {https://doi.org/10.18653/v1/2021.emnlp-main.126} {Improving zero-shot cross-lingual transfer learning via robust training}.
\newblock In \emph{Proceedings of the 2021 Conference on Empirical Methods in Natural Language Processing}, pages 1684--1697, Online and Punta Cana, Dominican Republic. Association for Computational Linguistics.

\bibitem[{Joshi et~al.(2020)Joshi, Santy, Budhiraja, Bali, and Choudhury}]{joshi-etal-2020-state}
Pratik Joshi, Sebastin Santy, Amar Budhiraja, Kalika Bali, and Monojit Choudhury. 2020.
\newblock \href {https://doi.org/10.18653/v1/2020.acl-main.560} {The state and fate of linguistic diversity and inclusion in the {NLP} world}.
\newblock In \emph{Proceedings of the 58th Annual Meeting of the Association for Computational Linguistics}, pages 6282--6293, Online. Association for Computational Linguistics.

\bibitem[{Kumar et~al.(2022)Kumar, Dandapat, and Choudhury}]{kumar-etal-2022-diversity}
Shanu Kumar, Sandipan Dandapat, and Monojit Choudhury. 2022.
\newblock \href {https://doi.org/10.18653/v1/2022.findings-naacl.78} {{''}diversity and uncertainty in moderation{''} are the key to data selection for multilingual few-shot transfer}.
\newblock In \emph{Findings of the Association for Computational Linguistics: NAACL 2022}, pages 1042--1055, Seattle, United States. Association for Computational Linguistics.

\bibitem[{Lauscher et~al.(2020)Lauscher, Ravishankar, Vuli{\'c}, and Glava{\v{s}}}]{lauscher-etal-2020-zero}
Anne Lauscher, Vinit Ravishankar, Ivan Vuli{\'c}, and Goran Glava{\v{s}}. 2020.
\newblock \href {https://doi.org/10.18653/v1/2020.emnlp-main.363} {From zero to hero: {O}n the limitations of zero-shot language transfer with multilingual {T}ransformers}.
\newblock In \emph{Proceedings of the 2020 Conference on Empirical Methods in Natural Language Processing (EMNLP)}, pages 4483--4499, Online. Association for Computational Linguistics.

\bibitem[{Lee et~al.(2019)Lee, Yue, and Li}]{lee19d_interspeech}
Grandee Lee, Xianghu Yue, and Haizhou Li. 2019.
\newblock \href {https://doi.org/10.21437/Interspeech.2019-1382} {{Linguistically Motivated Parallel Data Augmentation for Code-Switch Language Modeling}}.
\newblock In \emph{Proc. Interspeech 2019}, pages 3730--3734.

\bibitem[{Levesque et~al.(2012)Levesque, Davis, and Morgenstern}]{levesque2012winograd}
Hector Levesque, Ernest Davis, and Leora Morgenstern. 2012.
\newblock \href {https://cdn.aaai.org/ocs/4492/4492-21843-1-PB.pdf} {The winograd schema challenge}.
\newblock In \emph{Thirteenth international conference on the principles of knowledge representation and reasoning}.

\bibitem[{Lin et~al.(2022)Lin, Mihaylov, Artetxe, Wang, Chen, Simig, Ott, Goyal, Bhosale, Du, Pasunuru, Shleifer, Koura, Chaudhary, O{'}Horo, Wang, Zettlemoyer, Kozareva, Diab, Stoyanov, and Li}]{lin-etal-2022-shot}
Xi~Victoria Lin, Todor Mihaylov, Mikel Artetxe, Tianlu Wang, Shuohui Chen, Daniel Simig, Myle Ott, Naman Goyal, Shruti Bhosale, Jingfei Du, Ramakanth Pasunuru, Sam Shleifer, Punit~Singh Koura, Vishrav Chaudhary, Brian O{'}Horo, Jeff Wang, Luke Zettlemoyer, Zornitsa Kozareva, Mona Diab, Veselin Stoyanov, and Xian Li. 2022.
\newblock \href {https://aclanthology.org/2022.emnlp-main.616} {Few-shot learning with multilingual generative language models}.
\newblock In \emph{Proceedings of the 2022 Conference on Empirical Methods in Natural Language Processing}, pages 9019--9052, Abu Dhabi, United Arab Emirates. Association for Computational Linguistics.

\bibitem[{Lin et~al.(2019)Lin, Chen, Lee, Li, Zhang, Xia, Rijhwani, He, Zhang, Ma, Anastasopoulos, Littell, and Neubig}]{lin-etal-2019-choosing}
Yu-Hsiang Lin, Chian-Yu Chen, Jean Lee, Zirui Li, Yuyan Zhang, Mengzhou Xia, Shruti Rijhwani, Junxian He, Zhisong Zhang, Xuezhe Ma, Antonios Anastasopoulos, Patrick Littell, and Graham Neubig. 2019.
\newblock \href {https://doi.org/10.18653/v1/P19-1301} {Choosing transfer languages for cross-lingual learning}.
\newblock In \emph{Proceedings of the 57th Annual Meeting of the Association for Computational Linguistics}, pages 3125--3135, Florence, Italy. Association for Computational Linguistics.

\bibitem[{Liu et~al.(2020)Liu, Huang, Mungra, and Bowman}]{liu-etal-2020-precise}
Haokun Liu, William Huang, Dhara Mungra, and Samuel~R. Bowman. 2020.
\newblock \href {https://doi.org/10.18653/v1/2020.emnlp-main.664} {Precise task formalization matters in {W}inograd schema evaluations}.
\newblock In \emph{Proceedings of the 2020 Conference on Empirical Methods in Natural Language Processing (EMNLP)}, pages 8275--8280, Online. Association for Computational Linguistics.

\bibitem[{Mostafazadeh et~al.(2016)Mostafazadeh, Chambers, He, Parikh, Batra, Vanderwende, Kohli, and Allen}]{mostafazadeh-etal-2016-corpus}
Nasrin Mostafazadeh, Nathanael Chambers, Xiaodong He, Devi Parikh, Dhruv Batra, Lucy Vanderwende, Pushmeet Kohli, and James Allen. 2016.
\newblock \href {https://doi.org/10.18653/v1/N16-1098} {A corpus and cloze evaluation for deeper understanding of commonsense stories}.
\newblock In \emph{Proceedings of the 2016 Conference of the North {A}merican Chapter of the Association for Computational Linguistics: Human Language Technologies}, pages 839--849, San Diego, California. Association for Computational Linguistics.

\bibitem[{Muennighoff et~al.(2023)Muennighoff, Wang, Sutawika, Roberts, Biderman, Le~Scao, Bari, Shen, Yong, Schoelkopf, Tang, Radev, Aji, Almubarak, Albanie, Alyafeai, Webson, Raff, and Raffel}]{muennighoff2022crosslingual}
Niklas Muennighoff, Thomas Wang, Lintang Sutawika, Adam Roberts, Stella Biderman, Teven Le~Scao, M~Saiful Bari, Sheng Shen, Zheng~Xin Yong, Hailey Schoelkopf, Xiangru Tang, Dragomir Radev, Alham~Fikri Aji, Khalid Almubarak, Samuel Albanie, Zaid Alyafeai, Albert Webson, Edward Raff, and Colin Raffel. 2023.
\newblock \href {https://doi.org/10.18653/v1/2023.acl-long.891} {Crosslingual generalization through multitask finetuning}.
\newblock In \emph{Proceedings of the 61st Annual Meeting of the Association for Computational Linguistics (Volume 1: Long Papers)}, pages 15991--16111, Toronto, Canada. Association for Computational Linguistics.

\bibitem[{Nooralahzadeh et~al.(2020)Nooralahzadeh, Bekoulis, Bjerva, and Augenstein}]{nooralahzadeh-etal-2020-zero}
Farhad Nooralahzadeh, Giannis Bekoulis, Johannes Bjerva, and Isabelle Augenstein. 2020.
\newblock \href {https://doi.org/10.18653/v1/2020.emnlp-main.368} {Zero-shot cross-lingual transfer with meta learning}.
\newblock In \emph{Proceedings of the 2020 Conference on Empirical Methods in Natural Language Processing (EMNLP)}, pages 4547--4562, Online. Association for Computational Linguistics.

\bibitem[{Ponti et~al.(2020)Ponti, Glava{\v{s}}, Majewska, Liu, Vuli{\'c}, and Korhonen}]{ponti-etal-2020-xcopa}
Edoardo~Maria Ponti, Goran Glava{\v{s}}, Olga Majewska, Qianchu Liu, Ivan Vuli{\'c}, and Anna Korhonen. 2020.
\newblock \href {https://doi.org/10.18653/v1/2020.emnlp-main.185} {{XCOPA}: A multilingual dataset for causal commonsense reasoning}.
\newblock In \emph{Proceedings of the 2020 Conference on Empirical Methods in Natural Language Processing (EMNLP)}, pages 2362--2376, Online. Association for Computational Linguistics.

\bibitem[{Ponti et~al.(2019)Ponti, O{'}Horan, Berzak, Vuli{\'c}, Reichart, Poibeau, Shutova, and Korhonen}]{ponti-etal-2019-modeling}
Edoardo~Maria Ponti, Helen O{'}Horan, Yevgeni Berzak, Ivan Vuli{\'c}, Roi Reichart, Thierry Poibeau, Ekaterina Shutova, and Anna Korhonen. 2019.
\newblock \href {https://doi.org/10.1162/coli_a_00357} {Modeling language variation and universals: A survey on typological linguistics for natural language processing}.
\newblock \emph{Computational Linguistics}, 45(3):559--601.

\bibitem[{Pratapa et~al.(2018)Pratapa, Bhat, Choudhury, Sitaram, Dandapat, and Bali}]{pratapa-etal-2018-language}
Adithya Pratapa, Gayatri Bhat, Monojit Choudhury, Sunayana Sitaram, Sandipan Dandapat, and Kalika Bali. 2018.
\newblock \href {https://doi.org/10.18653/v1/P18-1143} {Language modeling for code-mixing: The role of linguistic theory based synthetic data}.
\newblock In \emph{Proceedings of the 56th Annual Meeting of the Association for Computational Linguistics (Volume 1: Long Papers)}, pages 1543--1553, Melbourne, Australia. Association for Computational Linguistics.

\bibitem[{Roemmele et~al.(2011)Roemmele, Bejan, and Gordon}]{roemmele2011choice}
Melissa Roemmele, Cosmin~Adrian Bejan, and Andrew~S Gordon. 2011.
\newblock Choice of plausible alternatives: An evaluation of commonsense causal reasoning.
\newblock In \emph{AAAI spring symposium: logical formalizations of commonsense reasoning}, pages 90--95.

\bibitem[{Scao et~al.(2022)Scao, Fan, Akiki, Pavlick, Ili{\'c}, Hesslow, Castagn{\'e}, Luccioni, Yvon, Gall{\'e} et~al.}]{scao2022bloom}
Teven~Le Scao, Angela Fan, Christopher Akiki, Ellie Pavlick, Suzana Ili{\'c}, Daniel Hesslow, Roman Castagn{\'e}, Alexandra~Sasha Luccioni, Fran{\c{c}}ois Yvon, Matthias Gall{\'e}, et~al. 2022.
\newblock \href {https://arxiv.org/abs/2211.05100} {Bloom: A 176b-parameter open-access multilingual language model}.
\newblock \emph{arXiv preprint arXiv:2211.05100}.

\bibitem[{Srinivasan et~al.(2022)Srinivasan, Kholkar, Kejriwal, Ganu, Dandapat, Sitaram, Santhanam, Aditya, Bali, and Choudhury}]{tool_2022}
Anirudh Srinivasan, Gauri Kholkar, Rahul Kejriwal, Tanuja Ganu, Sandipan Dandapat, Sunayana Sitaram, Balakrishnan Santhanam, Somak Aditya, Kalika Bali, and Monojit Choudhury. 2022.
\newblock \href {https://doi.org/10.1609/aaai.v36i11.21736} {Litmus predictor: An ai assistant for building reliable, high-performing and fair multilingual nlp systems}.
\newblock \emph{Proceedings of the AAAI Conference on Artificial Intelligence}, 36(11):13227--13229.

\bibitem[{Tarunesh et~al.(2021)Tarunesh, Kumar, and Jyothi}]{tarunesh-etal-2021-machine}
Ishan Tarunesh, Syamantak Kumar, and Preethi Jyothi. 2021.
\newblock \href {https://doi.org/10.18653/v1/2021.acl-long.245} {From machine translation to code-switching: Generating high-quality code-switched text}.
\newblock In \emph{Proceedings of the 59th Annual Meeting of the Association for Computational Linguistics and the 11th International Joint Conference on Natural Language Processing (Volume 1: Long Papers)}, pages 3154--3169, Online. Association for Computational Linguistics.

\bibitem[{Tikhonov and Ryabinin(2021)}]{tikhonov-ryabinin-2021-heads}
Alexey Tikhonov and Max Ryabinin. 2021.
\newblock \href {https://doi.org/10.18653/v1/2021.findings-acl.310} {{I}t{'}s {A}ll in the {H}eads: {U}sing {A}ttention {H}eads as a {B}aseline for {C}ross-{L}ingual {T}ransfer in {C}ommonsense {R}easoning}.
\newblock In \emph{Findings of the Association for Computational Linguistics: ACL-IJCNLP 2021}, pages 3534--3546, Online. Association for Computational Linguistics.

\bibitem[{Touvron et~al.(2023{\natexlab{a}})Touvron, Lavril, Izacard, Martinet, Lachaux, Lacroix, Rozi{\`e}re, Goyal, Hambro, Azhar et~al.}]{touvron2023llama}
Hugo Touvron, Thibaut Lavril, Gautier Izacard, Xavier Martinet, Marie-Anne Lachaux, Timoth{\'e}e Lacroix, Baptiste Rozi{\`e}re, Naman Goyal, Eric Hambro, Faisal Azhar, et~al. 2023{\natexlab{a}}.
\newblock \href {https://arxiv.org/abs/2302.13971} {Llama: Open and efficient foundation language models}.
\newblock \emph{arXiv preprint arXiv:2302.13971}.

\bibitem[{Touvron et~al.(2023{\natexlab{b}})Touvron, Martin, Stone, Albert, Almahairi, Babaei, Bashlykov, Batra, Bhargava, Bhosale et~al.}]{touvron2023llama2}
Hugo Touvron, Louis Martin, Kevin Stone, Peter Albert, Amjad Almahairi, Yasmine Babaei, Nikolay Bashlykov, Soumya Batra, Prajjwal Bhargava, Shruti Bhosale, et~al. 2023{\natexlab{b}}.
\newblock \href {https://arxiv.org/abs/2307.09288} {Llama 2: Open foundation and fine-tuned chat models}.
\newblock \emph{arXiv preprint arXiv:2307.09288}.

\bibitem[{Wang et~al.(2023)Wang, Kordi, Mishra, Liu, Smith, Khashabi, and Hajishirzi}]{wang-etal-2023-self-instruct}
Yizhong Wang, Yeganeh Kordi, Swaroop Mishra, Alisa Liu, Noah~A. Smith, Daniel Khashabi, and Hannaneh Hajishirzi. 2023.
\newblock \href {https://doi.org/10.18653/v1/2023.acl-long.754} {Self-instruct: Aligning language models with self-generated instructions}.
\newblock In \emph{Proceedings of the 61st Annual Meeting of the Association for Computational Linguistics (Volume 1: Long Papers)}, pages 13484--13508, Toronto, Canada. Association for Computational Linguistics.

\bibitem[{Whitehouse et~al.(2022)Whitehouse, Christopoulou, and Iacobacci}]{whitehouse-etal-2022-entitycs}
Chenxi Whitehouse, Fenia Christopoulou, and Ignacio Iacobacci. 2022.
\newblock \href {https://aclanthology.org/2022.findings-emnlp.499} {{E}ntity{CS}: Improving zero-shot cross-lingual transfer with entity-centric code switching}.
\newblock In \emph{Findings of the Association for Computational Linguistics: EMNLP 2022}, pages 6698--6714, Abu Dhabi, United Arab Emirates. Association for Computational Linguistics.

\bibitem[{Winata et~al.(2023)Winata, Aji, Cahyawijaya, Mahendra, Koto, Romadhony, Kurniawan, Moeljadi, Prasojo, Fung, Baldwin, Lau, Sennrich, and Ruder}]{winata-etal-2023-nusax}
Genta~Indra Winata, Alham~Fikri Aji, Samuel Cahyawijaya, Rahmad Mahendra, Fajri Koto, Ade Romadhony, Kemal Kurniawan, David Moeljadi, Radityo~Eko Prasojo, Pascale Fung, Timothy Baldwin, Jey~Han Lau, Rico Sennrich, and Sebastian Ruder. 2023.
\newblock \href {https://aclanthology.org/2023.eacl-main.57} {{N}usa{X}: Multilingual parallel sentiment dataset for 10 {I}ndonesian local languages}.
\newblock In \emph{Proceedings of the 17th Conference of the European Chapter of the Association for Computational Linguistics}, pages 815--834, Dubrovnik, Croatia. Association for Computational Linguistics.

\bibitem[{Winata et~al.(2019)Winata, Madotto, Wu, and Fung}]{winata-etal-2019-code}
Genta~Indra Winata, Andrea Madotto, Chien-Sheng Wu, and Pascale Fung. 2019.
\newblock \href {https://doi.org/10.18653/v1/K19-1026} {Code-switched language models using neural based synthetic data from parallel sentences}.
\newblock In \emph{Proceedings of the 23rd Conference on Computational Natural Language Learning (CoNLL)}, pages 271--280, Hong Kong, China. Association for Computational Linguistics.

\bibitem[{Wu et~al.(2023)Wu, Hu, Shi, Dziri, Suhr, Ammanabrolu, Smith, Ostendorf, and Hajishirzi}]{wu2023fine}
Zeqiu Wu, Yushi Hu, Weijia Shi, Nouha Dziri, Alane Suhr, Prithviraj Ammanabrolu, Noah~A Smith, Mari Ostendorf, and Hannaneh Hajishirzi. 2023.
\newblock \href {https://arxiv.org/abs/2306.01693} {Fine-grained human feedback gives better rewards for language model training}.
\newblock \emph{arXiv preprint arXiv:2306.01693}.

\bibitem[{Xia et~al.(2020)Xia, Anastasopoulos, Xu, Yang, and Neubig}]{xia-etal-2020-predicting}
Mengzhou Xia, Antonios Anastasopoulos, Ruochen Xu, Yiming Yang, and Graham Neubig. 2020.
\newblock \href {https://doi.org/10.18653/v1/2020.acl-main.764} {Predicting performance for natural language processing tasks}.
\newblock In \emph{Proceedings of the 58th Annual Meeting of the Association for Computational Linguistics}, pages 8625--8646, Online. Association for Computational Linguistics.

\bibitem[{Xue et~al.(2021)Xue, Constant, Roberts, Kale, Al-Rfou, Siddhant, Barua, and Raffel}]{xue-etal-2021-mt5}
Linting Xue, Noah Constant, Adam Roberts, Mihir Kale, Rami Al-Rfou, Aditya Siddhant, Aditya Barua, and Colin Raffel. 2021.
\newblock \href {https://doi.org/10.18653/v1/2021.naacl-main.41} {m{T}5: A massively multilingual pre-trained text-to-text transformer}.
\newblock In \emph{Proceedings of the 2021 Conference of the North American Chapter of the Association for Computational Linguistics: Human Language Technologies}, pages 483--498, Online. Association for Computational Linguistics.

\end{thebibliography}
\bibliographystyle{acl_natbib}

\appendix


        


\section{Model Details}
\label{sec:models_urls}

The open-source models used in the experiments are as follows:
\begin{itemize}
\item mBERT: \url{https://huggingface.co/bert-base-multilingual-uncased}
\item  XLMR-base: \url{https://huggingface.co/xlm-roberta-base}
\item  XLMR-large: \url{https://huggingface.co/xlm-roberta-large}
\item  Dolly-v2: \url{https://huggingface.co/databricks/dolly-v2-12b}
\item  StableVinuca: \url{https://huggingface.co/CarperAI/stable-vicuna-13b-delta}
\end{itemize}

\begin{figure*}

\centering
    \includegraphics[width=1\linewidth]{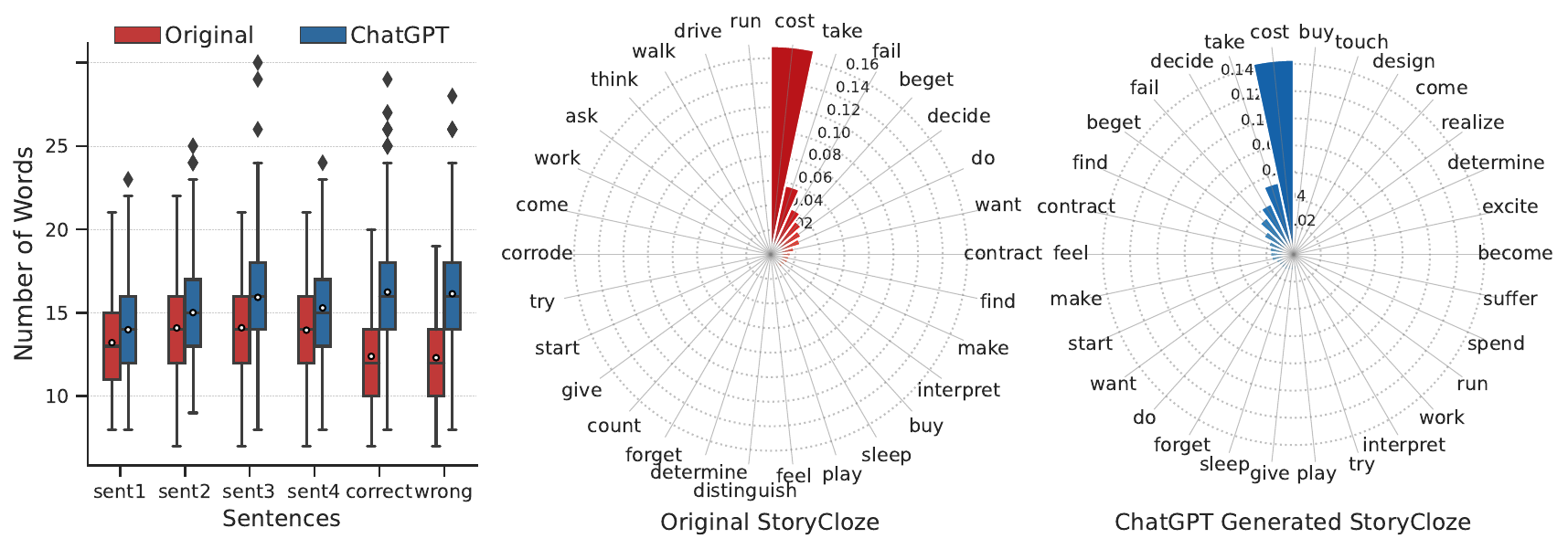}
\caption{Comparison between the 30 most frequent events and the lengths of the sentences in the original and the ChatGPT-generated English StoryCloze dataset.}
\label{fig:story}
\end{figure*}

\section{Sentences and Event Diversity of ChatGPT-generated StoryCloze Data}
\label{sec:diversity}
As the StoryCloze dataset contains more sentences and has richer content, we follow the analysis of the ROC story and further compare the stylistic features in terms of sentence length, and the most frequent events\footnote{Here we follow \citet{mostafazadeh-etal-2016-corpus} where an event is counted as any hyponym of ``event'' or ``process'' in WordNet.} generated by ChatGPT with the original data.
This helps us to determine whether ChatGPT-generated data can capture the corpus distribution by randomly sampling $n$ examples from the dataset in the instructions.

In \autoref{fig:story}, we present the results of comparing the generated data points with the original 300 train set used as few-shot examples in the generation instructions.
We can see that 23 of the 30 most frequent events in the original dataset can also be found in the 30 most frequent events of the ChatGPT-generated data.
Regarding the sentence length, we observe that ChatGPT tends to generate longer sentences, especially for the ending sentences, whereas in the original dataset, they tend to be the shortest among all sentences.

\section{Fixed Ratio Data Augmentation}
\label{sec:ratio}
We experiment with generating data with a fixed ratio of the original datasets. 
Specifically, we compare training with the original English data (200 randomly selected examples) and augment it with different quantities of English examples generated by GPT-4, where we include original training instances in all cases.

\begin{table}[!t]
\centering
\scalebox{0.8}{
\addtolength{\tabcolsep}{-2.5pt}
\begin{tabular}{ll|ccc}
    \toprule
\textbf{Model}                       & \textbf{Ratio} & \textbf{XCOPA} & \textbf{XWingrad} & \textbf{XStoryCloze} \\
\midrule
\multirow{4}{*}{mBERT}      
& 1$\times$    &  64.0    &  50.2  &  74.6         \\
 & 2$\times$   & 64.8     &  51.9 &    76.8      \\
& 5$\times$  &   68.0     &  57.1 &     80.6      \\
& 10$\times$  &   69.8    & 65.7 &  80.3     \\
 \midrule
\multirow{4}{*}{XLMR-Base}  
& 1$\times$    &  58.0  &    45.9   &  70.7     \\
& 2$\times$   &  59.0   &   53.7   &   79.7    \\
& 5$\times$  & 63.0  & 67.8 &    81.9      \\
& 10$\times$   &  65.8	   &  71.2   &  84.1    \\
 \midrule
\multirow{4}{*}{XLMR-Large} 
& 1$\times$  &  56.0   &     78.1   &  81.1     \\
& 2$\times$  &  61.2    &   79.8    &    90.9    \\
& 5$\times$  &  81.4     &    82.0      &   89.9   \\
& 10$\times$  &   85.2    &    82.8      &    91.9   \\
\bottomrule
\end{tabular}
}
\caption{Performance on English test examples training on GPT-4-generated English data and the original data. Original data points selected from the three datasets are set to 200. 1$\times$ corresponds to using only the original data, 2$\times$ means using 200 original data and 200 generated data. }
\label{tab:ratio}
\end{table}

The results in \autoref{tab:ratio} showcase the performance on English test examples when fine-tuning mBERT and XLMR models with training data sizes that are 1$\times$, 2$\times$, 5$\times$, and 10$\times$ the size of the original dataset.
We can see that performance consistently improves as we increase the amount of generated data except XStoryCloze, which has the highest baselines, echoing the previous findings. The relative performance gain is generally more pronounced when increasing the data from 2$\times$ to 5$\times$ for the other two datasets.

\section{Additional Results}
\label{sec:additional}

This section includes the following additional results:
\autoref{tab:all-copa}, \autoref{tab:xwino}, and \autoref{tab:xstory} show generated data in English with different LLMs on XCOPA, XWinograd, and XStoryCloze.
\autoref{tab:xcopa_full}  and \autoref{tab:gpt4-copa} show the full result on XCOPA with ChatGPT and GPT-4.


\begin{table*}[t]
\centering
\scalebox{0.8}{
\addtolength{\tabcolsep}{-2pt}
\begin{tabular}{lll|ccccccccccccc}
\toprule

\textbf{Fine-tuned} &
  \textbf{Train Data} &
  \textbf{LLM} &
  \textbf{AVG} &
  \textbf{EN} &
  \textbf{ET} &
  \textbf{HT} &
  \textbf{ID} &
  \textbf{IT} &
  \textbf{QU} &
  \textbf{SW} &
  \textbf{TA} &
  \textbf{TH} &
  \textbf{TR} &
  \textbf{VI} &
  \textbf{ZH} \\
  \midrule
\multirow{8}{*}{MBERT} &
  \multirow{4}{*}{GEN} &
  Dolly-v2 &
  54.0 &
  63.4 &
  52.0 &
  52.2 &
  54.0 &
  53.8 &
  47.6 &
  48.6 &
  53.4 &
  \textbf{53.4} &
  52.8 &
  50.4 &
  58.2 \\
 &
   &
  StableVicuna &
  53.5 &
  62.4 &
  51.6 &
  49.2 &
  55.8 &
  55.8 &
  50.0 &
  \textbf{50.2} &
  50.2 &
  52.6 &
  51.0 &
  50.4 &
  56.0 \\
 &
   &
  ChatGPT &
  56.0 &
  64.8 &
  54.8 &
  52.6 &
  58.0 &
  57.4 &
  49.8 &
  48.4 &
  \textbf{55.6} &
  52.8 &
  53.2 &
  53.0 &
  59.0 \\
 &
   &
  GPT-4 &
  \textbf{58.2} &
  \textbf{69.2} &
  \textbf{59.2} &
  \textbf{54.0} &
  \textbf{60.6} &
  \textbf{59.2} &
  \textbf{50.8} &
  48.2 &
  55.0 &
  48.2 &
  \textbf{53.8} &
  \textbf{57.6} &
  \textbf{61.0} \\
    \cmidrule(l){2-16} 
 &
  \multirow{4}{*}{GEN+ORI} &
  Dolly-v2 &
  54.4 &
  59.8 &
  52.6 &
  53.2 &
  53.0 &
  56.4 &
  \textbf{53.8} &
  \textbf{52.4} &
  50.4 &
  \textbf{54.8} &
  49.8 &
  52.6 &
  58.8 \\
 &
   &
  StableVicuna &
  55.6 &
  65.2 &
  53.4 &
  50.4 &
  59.0 &
  60.0 &
  51.6 &
  50.4 &
  49.4 &
  52.0 &
  52.4 &
  54.0 &
  58.2 \\
 &
   &
  ChatGPT &
  54.6 &
  59.6 &
  56.4 &
  \textbf{53.6} &
  53.8 &
  51.4 &
  51.4 &
  51.6 &
  \textbf{50.4} &
  52.6 &
  \textbf{54.0} &
  55.0 &
  59.2 \\
 &
   &
  GPT-4 &
  \textbf{59.3} &
  \textbf{72.6} &
  \textbf{58.8} &
  53.0 &
  \textbf{62.0} &
  \textbf{61.0} &
  53.0 &
  50.0 &
  54.0 &
  48.2 &
  52.0 &
  \textbf{57.6} &
  \textbf{64.6} \\
    \cmidrule{1-16} 
\multirow{8}{*}{XLMR-Base} &
  \multirow{4}{*}{GEN} &
  Dolly-v2 &
  59.0 &
  64.4 &
  58.8 &
  52.8 &
  60.8 &
  61.0 &
  50.8 &
  55.6 &
  60.4 &
  58.0 &
  57.2 &
  58.6 &
  59.0 \\
 &
   &
  StableVicuna &
  58.5 &
  60.4 &
  59.4 &
  \textbf{53.6} &
  60.8 &
  56.8 &
  49.2 &
  56.0 &
  61.2 &
  60.4 &
  54.8 &
  59.6 &
  58.6 \\
 &
   &
  ChatGPT &
  58.8 &
  62.4 &
  56.4 &
  52.4 &
  61.4 &
  58.6 &
  \textbf{52.2} &
  52.0 &
  63.4 &
  61.2 &
  56.4 &
  59.6 &
  62.8 \\
 &
   &
  GPT-4 &
  \textbf{63.6} &
  \textbf{67.0} &
  \textbf{62.4} &
  52.0 &
  \textbf{68.6} &
  \textbf{62.6} &
  51.8 &
  \textbf{58.6} &
  \textbf{65.4} &
  \textbf{64.8} &
  \textbf{63.2} &
  \textbf{66.6} &
  \textbf{69.6} \\
   \cmidrule(l){2-16} 
 &
  \multirow{4}{*}{GEN+ORI} &
  Dolly-v2 &
  58.7 &
  65.6 &
  57.6 &
  \textbf{52.2} &
  60.8 &
  58.4 &
  52.4 &
  58.2 &
  57.4 &
  58.0 &
  58.4 &
  58.0 &
  59.8 \\
 &
   &
  StableVicuna &
  61.1 &
  65.0 &
  62.4 &
  49.4 &
  64.2 &
  62.4 &
  46.2 &
  \textbf{60.4} &
  59.6 &
  58.0 &
  58.0 &
  63.0 &
  63.4 \\
 &
   &
  ChatGPT &
  59.8 &
  63.8 &
  61.6 &
  51.6 &
  62.6 &
  59.8 &
  51.2 &
  51.6 &
  60.4 &
  61.6 &
  61.8 &
  64.8 &
  62.0 \\
 &
   &
  GPT-4 &
  \textbf{63.6} &
  \textbf{69.6} &
  \textbf{63.8} &
  51.2 &
  \textbf{67.2} &
  \textbf{62.4} &
  \textbf{52.6} &
  58.4 &
  \textbf{63.8} &
  \textbf{66.0} &
  \textbf{64.2} &
  \textbf{66.8} &
  \textbf{69.4} \\
   \cmidrule{1-16} 
\multirow{8}{*}{XLMR-Large} &
  \multirow{4}{*}{GEN} &
  Dolly-v2 &
  59.6 &
  62.4 &
  58.6 &
  49.6 &
  64.8 &
  59.2 &
  50.6 &
  56.8 &
  60.8 &
  58.8 &
  57.0 &
  61.0 &
  63.0 \\
 &
   &
  StableVicuna &
  65.7 &
  71.4 &
  66.2 &
  50.4 &
  71.4 &
  70.2 &
  50.0 &
  60.0 &
  64.0 &
  63.6 &
  68.0 &
  68.2 &
  69.8 \\
 &
   &
  ChatGPT &
  65.2 &
  71.2 &
  64.6 &
  51.6 &
  70.8 &
  66.6 &
  \textbf{51.0} &
  58.8 &
  66.0 &
  68.2 &
  69.0 &
  68.8 &
  68.8 \\
 &
   &
  GPT-4 &
  \textbf{73.6} &
  \textbf{83.2} &
  \textbf{71.2} &
  \textbf{52.0} &
  \textbf{81.2} &
  \textbf{78.2} &
  \textbf{51.0} &
  \textbf{62.2} &
  \textbf{76.6} &
  \textbf{77.4} &
  \textbf{75.0} &
  \textbf{78.4} &
  \textbf{79.0} \\

  \cmidrule(l){2-16} 
 &
  \multirow{4}{*}{GEN+ORI} &
  Dolly-v2 &
  66.4 &
  74.2 &
  62.8 &
  \textbf{53.0} &
  72.0 &
  70.4 &
  46.2 &
  61.6 &
  65.6 &
  66.2 &
  69.6 &
  67.6 &
  70.6 \\
 &
   &
  StableVicuna &
  69.9 &
  76.0 &
  69.8 &
  51.2 &
  75.0 &
  74.2 &
  51.2 &
  64.4 &
  70.2 &
  71.6 &
  72.2 &
  72.6 &
  75.4 \\
 &
   &
  ChatGPT &
  69.5 &
  76.4 &
  69.8 &
  48.2 &
  76.0 &
  72.8 &
  50.8 &
  63.4 &
  67.8 &
  70.8 &
  70.2 &
  73.4 &
  77.8 \\
 &
   &
  GPT-4 &
  \textbf{73.7} &
  \textbf{84.6} &
  \textbf{70.4} &
  50.0 &
  \textbf{80.8} &
  \textbf{80.2} &
  \textbf{51.8} &
  \textbf{65.8} &
  \textbf{72.8} &
  \textbf{76.0} &
  \textbf{74.8} &
  \textbf{78.4} &
  \textbf{80.4}
\\
\bottomrule
\end{tabular}}
\caption{Accuracy on XCOPA with English generated data from different LLMs.
}
 \label{tab:all-copa}
\end{table*}
\begin{table*}[!ht]
\centering
\scalebox{0.85}{
\addtolength{\tabcolsep}{0pt}
\begin{tabular}{lll|rrrrrrr}
\toprule

\textbf{Fine-tuned} &
  \textbf{Training data} &
  \textbf{LLM} &
  \multicolumn{1}{l}{\textbf{AVG}} &
  \multicolumn{1}{l}{\textbf{EN}} &
  \multicolumn{1}{l}{\textbf{FR}} &
  \multicolumn{1}{l}{\textbf{JA}} &
  \multicolumn{1}{l}{\textbf{PT}} &
  \multicolumn{1}{l}{\textbf{RU}} &
  \multicolumn{1}{l}{\textbf{ZH}} \\
  \midrule
\multirow{8}{*}{MBERT} &
  \multirow{4}{*}{GEN} &
  Dolly-v2 &
  \textbf{56.47} &
  \textbf{71.24} &
  53.01 &
  52.45 &
  \textbf{53.23} &
  54.92 &
  53.97 \\
 &
   &
  StableVicuna &
  53.73 &
  54.94 &
  \textbf{56.63} &
  50.26 &
  50.57 &
  52.06 &
  57.94 \\
 &
   &
  ChatGPT &
  56.00 &
  54.94 &
  54.22 &
  \textbf{54.01} &
  52.09 &
  \textbf{55.87} &
  \textbf{64.88} \\
 &
   &
  GPT-4 &
  54.90 &
  56.22 &
  \textbf{56.63} &
  52.55 &
  51.71 &
  52.38 &
  59.92 \\
  \cmidrule(l){2-10}
 &
  \multirow{4}{*}{GEN+ORI} &
  Dolly-v2 &
  \textbf{59.32} &
  \textbf{71.24} &
  57.83 &
  \textbf{53.81} &
  56.65 &
  59.05 &
  57.34 \\
 &
   &
  StableVicuna &
  58.46 &
  57.94 &
  63.86 &
  \textbf{53.81} &
  \textbf{57.41} &
  58.41 &
  59.33 \\
 &
   &
  ChatGPT &
  58.26 &
  56.65 &
  \textbf{66.27} &
  53.60 &
  56.27 &
  \textbf{60.00} &
  56.75 \\
 &
   &
  GPT-4 &
  57.48 &
  53.65 &
  62.65 &
  \textbf{54.43} &
  55.89 &
  57.14 &
  \textbf{61.11} \\
   \cmidrule{1-10}
\multirow{8}{*}{XLMR-Base} &
  \multirow{4}{*}{GEN} &
  Dolly-v2 &
  59.63 &
  \textbf{71.24} &
  57.83 &
  55.79 &
  \textbf{57.03} &
  57.78 &
  58.13 \\
 &
   &
  StableVicuna &
  58.95 &
  60.09 &
  55.42 &
  57.35 &
  52.47 &
  58.73 &
  69.64 \\
 &
   &
  ChatGPT &
  62.69 &
  69.10 &
  60.24 &
  61.42 &
  \textbf{57.03} &
  \textbf{61.27} &
  67.06 \\
 &
   &
  GPT-4 &
  \textbf{63.32} &
  69.10 &
  \textbf{61.45} &
  \textbf{61.52} &
  56.65 &
  60.95 &
  \textbf{70.24} \\
   \cmidrule(l){2-10}
 &
  \multirow{4}{*}{GEN+ORI} &
  Dolly-v2 &
  66.33 &
  \textbf{75.54} &
  63.86 &
  65.80 &
  \textbf{64.26} &
  62.86 &
  65.67 \\
 &
   &
  StableVicuna &
  65.97 &
  64.38 &
  66.27 &
  67.15 &
  63.88 &
  \textbf{65.71} &
  68.45 \\
 &
   &
  ChatGPT &
  65.94 &
  65.24 &
  60.24 &
  \textbf{68.93} &
  70.72 &
  62.86 &
  67.66 \\
   \cmidrule{1-10}
 &
   &
  GPT-4 &
  \textbf{66.88} &
  68.24 &
  \textbf{67.47} &
  66.94 &
  63.88 &
  63.49 &
  \textbf{71.23} \\
\multirow{8}{*}{XLMR-Large} &
  \multirow{4}{*}{GEN} &
  Dolly-v2 &
  \textbf{76.86} &
  \textbf{87.55} &
  67.47 &
  \textbf{81.02} &
  \textbf{76.43} &
  74.29 &
  74.40 \\
 &
   &
  StableVicuna &
  68.22 &
  74.25 &
  63.86 &
  68.20 &
  66.16 &
  63.81 &
  73.02 \\
 &
   &
  ChatGPT &
  73.20 &
  81.97 &
  66.27 &
  73.10 &
  66.92 &
  72.38 &
  78.57 \\
 &
   &
  GPT-4 &
  76.37 &
  81.55 &
  \textbf{74.70} &
  75.91 &
  71.86 &
  \textbf{75.24} &
  \textbf{78.97} \\
   \cmidrule(l){2-10}
 &
  \multirow{4}{*}{GEN+ORI} &
  Dolly-v2 &
  83.10 &
  \textbf{90.56} &
  79.52 &
  85.19 &
  84.03 &
  80.95 &
  78.37 \\
 &
   &
  StableVicuna &
  82.02 &
  83.26 &
  80.72 &
  83.84 &
  86.31 &
  \textbf{82.22} &
  75.79 \\
 &
   &
  ChatGPT &
  83.22 &
  85.84 &
  80.72 &
  \textbf{87.38} &
  85.93 &
  80.95 &
  78.50 \\
 &
   &
  GPT-4 &
  \textbf{83.52} &
  85.41 &
  \textbf{81.93} &
  85.92 &
  \textbf{86.69} &
  80.63 &
  \textbf{80.56}\\
  \bottomrule
\end{tabular}}
\caption{Accuracy on XWinograd with English generated data from different LLMs.
}
 \label{tab:xwino}
\end{table*}
\begin{table*}[!ht]
\centering
\scalebox{0.78}{
\addtolength{\tabcolsep}{-1pt}
\begin{tabular}{lll|rrrrrrrrrrrr}
\toprule

\textbf{Fine-tuned} &
  \textbf{Training data} &
  \textbf{LLM} &
  \textbf{AVG} &
  \textbf{EN} &
  \textbf{RU} &
  \textbf{ZH} &
  \textbf{ES} &
  \textbf{AR} &
  \textbf{HI} &
  \textbf{ID} &
  \textbf{TE} &
  \textbf{SW} &
  \textbf{EU} &
  \textbf{MY} \\
  \midrule
 &
   &
  Dolly-v2 &
  \textbf{68.7} &
  \textbf{78.8} &
  \textbf{71.3} &
  \textbf{73.6} &
  \textbf{74.2} &
  67.4 &
  66.9 &
  69.0 &
  \textbf{65.0} &
  60.9 &
  \textbf{66.8} &
  62.0 \\
 &
   &
  StableVicuna &
  64.6 &
  71.4 &
  66.8 &
  68.8 &
  68.1 &
  64.3 &
  63.6 &
  66.1 &
  61.2 &
  58.6 &
  63.6 &
  58.4 \\
 &
   &
  ChatGPT &
  64.3 &
  69.7 &
  66.4 &
  68.1 &
  68.0 &
  64.6 &
  64.5 &
  66.6 &
  59.8 &
  59.2 &
  62.3 &
  58.4 \\
 &
  \multirow{-4}{*}{GEN} &
  GPT-4 &
  68.0 &
  75.5 &
  70.8 &
  73.3 &
  70.4 &
  \textbf{67.6} &
  \textbf{68.2} &
  \textbf{69.6} &
  63.1 &
  \textbf{62.3} &
  65.4 &
  \textbf{62.2} \\
  \cmidrule(l){2-15}
 &
   &
  Dolly-v2 &
  68.1 &
  75.7 &
  71.2 &
  72.4 &
  73.2 &
  66.4 &
  67.1 &
  68.9 &
  \textbf{64.5} &
  61.4 &
  67.1 &
  61.0 \\
 &
   &
  StableVicuna &
  67.3 &
  77.0 &
  71.0 &
  70.2 &
  71.4 &
  67.2 &
  66.5 &
  68.4 &
  62.4 &
  60.5 &
  64.3 &
  61.4 \\
 &
   &
  ChatGPT &
  68.3 &
  76.4 &
  68.5 &
  72.9 &
  73.0 &
  66.3 &
  68.6 &
  71.1 &
  62.0 &
  \textbf{62.0} &
  67.4 &
  \textbf{63.4} \\
\multirow{-8}{*}{MBERT} &
  \multirow{-4}{*}{GEN+ORI} &
  GPT-4 &
  \textbf{69.8} &
  \textbf{79.5} &
  \textbf{73.1} &
  \textbf{75.3} &
  \textbf{73.4} &
  \textbf{68.1} &
  \textbf{69.8} &
  \textbf{71.9} &
  64.1 &
  \textbf{62.0} &
  \textbf{68.9} &
  61.6 \\
   \cmidrule{1-15}
 &
   &
  Dolly-v2 &
  75.8 &
  81.4 &
  79.2 &
  80.3 &
  78.0 &
  73.6 &
  74.7 &
  80.7 &
  73.0 &
  68.8 &
  72.2 &
  71.7 \\
 &
   &
  StableVicuna &
  69.6 &
  72.3 &
  71.1 &
  71.5 &
  70.4 &
  68.3 &
  70.4 &
  72.1 &
  68.4 &
  65.7 &
  68.0 &
  67.7 \\
 &
   &
  ChatGPT &
  67.4 &
  69.7 &
  68.9 &
  68.5 &
  68.7 &
  66.1 &
  68.2 &
  68.7 &
  67.0 &
  63.7 &
  65.6 &
  66.6 \\
 &
  \multirow{-4}{*}{GEN} &
  GPT-4 &
  \textbf{74.6} &
  \textbf{78.2} &
  \textbf{78.0} &
  \textbf{78.1} &
  \textbf{77.0} &
  \textbf{73.5} &
  \textbf{75.7} &
  \textbf{77.6} &
  \textbf{71.7} &
  \textbf{68.4} &
  \textbf{73.6} &
  \textbf{69.2} \\
   \cmidrule(l){2-15}
 &
   &
  Dolly-v2 &
  76.5 &
  81.5 &
  80.0 &
  80.5 &
  79.4 &
  75.1 &
  75.0 &
  79.6 &
  74.5 &
  71.5 &
  72.3 &
  72.6 \\
 &
   &
  StableVicuna &
  74.2 &
  79.2 &
  77.4 &
  77.8 &
  76.4 &
  74.0 &
  74.5 &
  78.2 &
  70.2 &
  67.6 &
  71.7 &
  69.6 \\
 &
   &
  ChatGPT &
  74.5 &
  78.0 &
  76.6 &
  78.8 &
  76.2 &
  72.9 &
  73.9 &
  78.9 &
  71.5 &
  69.6 &
  72.3 &
  71.0 \\
\multirow{-8}{*}{XLMR-Base} &
  \multirow{-4}{*}{GEN+ORI} &
  GPT-4 &
  \textbf{79.3} &
  \textbf{85.4} &
  \textbf{83.2} &
  \textbf{82.6} &
  \textbf{83.0} &
  \textbf{78.0} &
  \textbf{79.9} &
  \textbf{82.7} &
  \textbf{75.9} &
  \textbf{72.9} &
  \textbf{74.9} &
  \textbf{74.3} \\
   \cmidrule{1-15}
 &
   &
  Dolly-v2 &
  84.8 &
  87.4 &
  87.3 &
  87.8 &
  86.6 &
  83.0 &
  84.4 &
  87.1 &
  \textbf{84.1} &
  81.0 &
  82.9 &
  81.4 \\
 &
   &
  StableVicuna &
  74.6 &
  76.7 &
  75.9 &
  77.4 &
  76.2 &
  72.9 &
  74.5 &
  76.2 &
  74.3 &
  70.8 &
  73.5 &
  72.5 \\
 &
   &
  ChatGPT &
  77.3 &
  78.6 &
  79.9 &
  78.0 &
  77.9 &
  75.8 &
  77.4 &
  78.0 &
  76.4 &
  73.5 &
  77.1 &
  77.7 \\
 &
  \multirow{-4}{*}{GEN} &
  GPT-4 &
  \textbf{86.0} &
  \textbf{88.5} &
  \textbf{88.2} &
  \textbf{88.2} &
  \textbf{88.0} &
  \textbf{84.9} &
  \textbf{85.7} &
  \textbf{87.8} &
  83.7 &
  \textbf{81.3} &
  \textbf{85.6} &
  \textbf{84.3} \\
   \cmidrule(l){2-15}
 &
 
   &
  Dolly-v2 &
  86.4 &
  89.2 &
  87.2 &
  89.5 &
  87.1 &
  85.2 &
  86.7 &
  87.7 &
  \textbf{85.0} &
  83.0 &
  85.7 &
  83.8 \\
 &
   &
  StableVicuna &
  84.8 &
  88.4 &
  87.6 &
  87.8 &
  86.6 &
  82.9 &
  83.3 &
  87.4 &
  83.7 &
  81.3 &
  83.7 &
  80.0 \\
 &
   &
  ChatGPT &
  85.8 &
  88.5 &
  88.0 &
  88.3 &
  87.3 &
  83.7 &
  85.9 &
  87.2 &
  83.7 &
  81.6 &
  85.4 &
  83.8 \\
\multirow{-8}{*}{XLMR-Large} &
  \multirow{-4}{*}{GEN+ORI} &
  GPT-4 &
  \cellcolor[HTML]{FFFFFF}\textbf{88.4} &
  \textbf{92.3} &
  \textbf{91.5} &
  \textbf{91.5} &
  \textbf{90.5} &
  \textbf{86.4} &
  \textbf{88.4} &
  \textbf{91.1} &
  84.8 &
  \textbf{83.1} &
  \textbf{87.4} &
  \textbf{85.2} \\
  \bottomrule
\end{tabular}}
\caption{Accuracy on XStoryCloze with English generated data from different LLMs.
}
 \label{tab:xstory}
\end{table*}

\begin{table*}[!ht]
\centering
\scalebox{0.75}{
\addtolength{\tabcolsep}{-1.5pt}
\begin{tabular}{llr|ccccccccccccc}
\toprule
\multicolumn{1}{l}{\cellcolor[HTML]{FFFFFF}\textbf{Model}} &
  \multicolumn{1}{l}{\cellcolor[HTML]{FFFFFF}\textbf{Training Data}} &
  \multicolumn{1}{r|}{\cellcolor[HTML]{FFFFFF}\textbf{|Data|}} &
  \multicolumn{1}{c}{\cellcolor[HTML]{FFFFFF}\textbf{AVG}} &
  \multicolumn{1}{c}{\cellcolor[HTML]{FFFFFF}\textbf{EN}} &
  \multicolumn{1}{c}{\cellcolor[HTML]{FFFFFF}\textbf{ET}} &
  \multicolumn{1}{c}{\cellcolor[HTML]{FFFFFF}\textbf{HT}} &
  \multicolumn{1}{c}{\cellcolor[HTML]{FFFFFF}\textbf{ID}} &
  \multicolumn{1}{c}{\cellcolor[HTML]{FFFFFF}\textbf{IT}} &
  \multicolumn{1}{c}{\cellcolor[HTML]{FFFFFF}\textbf{QU}} &
  \multicolumn{1}{c}{\cellcolor[HTML]{FFFFFF}\textbf{SW}} &
  \multicolumn{1}{c}{\cellcolor[HTML]{FFFFFF}\textbf{TA}} &
  \multicolumn{1}{c}{\cellcolor[HTML]{FFFFFF}\textbf{TH}} &
  \multicolumn{1}{c}{\cellcolor[HTML]{FFFFFF}\textbf{TR}} &
  \multicolumn{1}{c}{\cellcolor[HTML]{FFFFFF}\textbf{VI}} &
  \multicolumn{1}{c}{\cellcolor[HTML]{FFFFFF}\textbf{ZH}} \\\midrule
\cellcolor[HTML]{FFFFFF} &
 \small{$ORI$ \sc({baseline)}}  &
  400 &
47.2
& 53.8 & 44.2 & 48.6 & 47.2 & 46.2 & 50.6 & 45.4 & 48.4 & 49.8 & 49.8 & 43.6 & 47.4 \\
\cellcolor[HTML]{FFFFFF} &
  \small{$GEN_{EN}$ } &
  3.7k &
  \cellcolor[HTML]{96D5B4}56.0 &
  \cellcolor[HTML]{8ACFAA}64.8 &
  \cellcolor[HTML]{8CD0AC}54.8 &
  \cellcolor[HTML]{BCE4D1}52.6 &
  \cellcolor[HTML]{8BD0AB}58.0 &
  \cellcolor[HTML]{89CFA9}57.4 &
  \cellcolor[HTML]{FBEDEC}49.8 &
  \cellcolor[HTML]{CDEBDD}48.4 &
  \cellcolor[HTML]{9FD8BB}55.6 &
  \cellcolor[HTML]{CDEBDD}52.8 &
  \cellcolor[HTML]{C6E8D8}53.2 &
  \cellcolor[HTML]{93D3B2}53.0 &
  \cellcolor[HTML]{87CEA7}59.0 \\
\cellcolor[HTML]{FFFFFF} &
  \small{$GEN_{EN}$  + $ORI$}  &
  4.1k &
  \cellcolor[HTML]{9ED8BB}54.6 &
  \cellcolor[HTML]{A7DCC2}59.6 &
  \cellcolor[HTML]{84CCA5}56.4 &
  \cellcolor[HTML]{ABDDC5}53.6 &
  \cellcolor[HTML]{A3DABE}53.8 &
  \cellcolor[HTML]{AADDC5}51.4 &
  \cellcolor[HTML]{F2FAF6}51.4 &
  \cellcolor[HTML]{A5DBC0}51.6 &
  \cellcolor[HTML]{DEF2E8}50.4 &
  \cellcolor[HTML]{D0ECDF}52.6 &
  \cellcolor[HTML]{B9E3CF}54.0 &
  \cellcolor[HTML]{88CEA8}55.0 &
  \cellcolor[HTML]{86CDA7}59.2 \\
\cellcolor[HTML]{FFFFFF} &
  \small{$GEN_{EN}$ + $ORI$ \sc{(tlv)}}  &
  4.1k &
  \cellcolor[HTML]{8ED1AD}57.6 &
  \cellcolor[HTML]{78C89C}68.0 &
  \cellcolor[HTML]{89CFA9}55.4 &
  \cellcolor[HTML]{A9DDC4}54.0 &
  \cellcolor[HTML]{7AC89D}61.2 &
  \cellcolor[HTML]{7CC99E}59.8 &
  \cellcolor[HTML]{EBF7F2}51.8 &
  \cellcolor[HTML]{A7DCC2}51.2 &
  \cellcolor[HTML]{9ED8BB}55.8 &
  \cellcolor[HTML]{B2E0CA}54.4 &
  \cellcolor[HTML]{D7EFE4}52.2 &
  \cellcolor[HTML]{91D2B0}53.4 &
  \cellcolor[HTML]{86CDA7}59.2 \\
\cellcolor[HTML]{FFFFFF} &
  \small{$GEN_{EN}$ } &
  28.6k &
  \cellcolor[HTML]{90D2AF}57.2 &
  \cellcolor[HTML]{82CCA4}66.2 &
  \cellcolor[HTML]{87CEA8}55.8 &
  \cellcolor[HTML]{DBF1E6}50.8 &
  \cellcolor[HTML]{88CEA8}58.6 &
  \cellcolor[HTML]{85CDA6}58.2 &
  \cellcolor[HTML]{D4EEE1}53.2 &
  \cellcolor[HTML]{A7DCC2}51.2 &
  \cellcolor[HTML]{96D5B4}57.2 &
  \cellcolor[HTML]{C6E8D8}53.2 &
  \cellcolor[HTML]{DBF1E6}52.0 &
  \cellcolor[HTML]{82CCA4}56.0 &
  \cellcolor[HTML]{7CC99E}61.0 \\
\cellcolor[HTML]{FFFFFF} &
  \small{$GEN_{EN}$  + $ORI$}  &
  29k &
  \cellcolor[HTML]{91D2B0}57.0 &
  \cellcolor[HTML]{80CBA2}66.6 &
  \cellcolor[HTML]{89CFA9}55.4 &
  \cellcolor[HTML]{D0ECDF}51.4 &
  \cellcolor[HTML]{85CDA6}59.2 &
  \cellcolor[HTML]{82CCA4}58.6 &
  \cellcolor[HTML]{E1F3EB}52.4 &
  \cellcolor[HTML]{A9DDC4}50.8 &
  \cellcolor[HTML]{AADDC5}53.6 &
  \cellcolor[HTML]{C6E8D8}53.2 &
  \cellcolor[HTML]{FCFEFD}50.0 &
  \cellcolor[HTML]{89CFA9}54.8 &
  \cellcolor[HTML]{72C596}62.8 \\
\cellcolor[HTML]{FFFFFF} &
  \small{$GEN_{EN}$ + $ORI$ \sc{(tlv)}}  &
29k &
  \cellcolor[HTML]{91D2B0}57.0 &
  \cellcolor[HTML]{80CBA2}66.6 &
  \cellcolor[HTML]{89CFA9}55.4 &
  \cellcolor[HTML]{D0ECDF}51.4 &
  \cellcolor[HTML]{85CDA6}59.2 &
  \cellcolor[HTML]{82CCA4}58.6 &
  \cellcolor[HTML]{E1F3EB}52.4 &
  \cellcolor[HTML]{A9DDC4}50.8 &
  \cellcolor[HTML]{AADDC5}53.6 &
  \cellcolor[HTML]{C6E8D8}53.2 &
  \cellcolor[HTML]{FCFEFD}50.0 &
  \cellcolor[HTML]{89CFA9}54.8 &
  \cellcolor[HTML]{72C596}62.8 \\
\cellcolor[HTML]{FFFFFF} &
  \small{$GEN_{XX}$ }&
  3.6k/lang &
  \cellcolor[HTML]{8ED1AD}57.5 &
  \cellcolor[HTML]{8ACFAA}64.8 &
  \cellcolor[HTML]{7CC99E}57.8 &
  \cellcolor[HTML]{96D5B4}57.4 &
  \cellcolor[HTML]{8BD0AB}58.0 &
  \cellcolor[HTML]{7AC89D}60.2 &
  \cellcolor[HTML]{BCE4D1}54.6 &
  \cellcolor[HTML]{A6DBC1}51.4 &
  \cellcolor[HTML]{B2E0CA}53.0 &
  \cellcolor[HTML]{FFFFFF}-- &
  \cellcolor[HTML]{FFFFFF}-- &
  \cellcolor[HTML]{93D3B2}53.0 &
  \cellcolor[HTML]{76C79A}62.0 \\
\cellcolor[HTML]{FFFFFF} &
  \small{\small{$GEN_{XX}$ } + $ORI$} &
4k &
  \cellcolor[HTML]{92D3B1}56.8 &
  \cellcolor[HTML]{A7DCC2}59.6 &
  \cellcolor[HTML]{76C79A}58.8 &
  \cellcolor[HTML]{A6DBC1}54.6 &
  \cellcolor[HTML]{95D4B3}56.2 &
  \cellcolor[HTML]{74C698}61.2 &
  \cellcolor[HTML]{CDEBDD}53.6 &
  \cellcolor[HTML]{94D4B2}54.6 &
  \cellcolor[HTML]{AADDC5}53.6 &
  \cellcolor[HTML]{FFFFFF}-- &
  \cellcolor[HTML]{FFFFFF}-- &
  \cellcolor[HTML]{99D5B6}52.0 &
  \cellcolor[HTML]{80CBA2}60.2 \\
\cellcolor[HTML]{FFFFFF} &
  \small{$GEN_{EN}^{Trans} + ORI$} &
4k &
  \cellcolor[HTML]{87CEA8}58.7 &
  \cellcolor[HTML]{A7DCC2}59.6 &
  \cellcolor[HTML]{71C495}59.8 &
  \cellcolor[HTML]{89CFA9}59.8 &
  \cellcolor[HTML]{71C495}62.8 &
  \cellcolor[HTML]{75C699}61.0 &
  \cellcolor[HTML]{FFFFFF}-- &
  \cellcolor[HTML]{9FD8BB}52.6 &
  \cellcolor[HTML]{99D5B6}56.8 &
  \cellcolor[HTML]{C3E7D6}53.4 &
  \cellcolor[HTML]{A4DABF}56.2 &
  \cellcolor[HTML]{76C79A}58.2 &
  \cellcolor[HTML]{85CDA6}59.4 \\
\multirow{-11}{*}{\cellcolor[HTML]{FFFFFF}\small{MBERT}} &
  \small{$GEN_{EN}^{Trans} + ORI$} &
  29k/lang &
  \cellcolor[HTML]{7DCA9F}\textbf{60.6} &
  \cellcolor[HTML]{80CBA2}66.6 &
  \cellcolor[HTML]{6EC393}61.8 &
  \cellcolor[HTML]{94D4B2}57.8 &
  \cellcolor[HTML]{7CC99E}60.8 &
  \cellcolor[HTML]{6EC393}62.2 &
  \cellcolor[HTML]{FFFFFF}-- &
  \cellcolor[HTML]{9CD7B9}53.2 &
  \cellcolor[HTML]{90D2AF}58.4 &
  \cellcolor[HTML]{C6E8D8}53.2 &
  \cellcolor[HTML]{7ECAA0}63.0 &
  \cellcolor[HTML]{6EC393}60.6 &
  \cellcolor[HTML]{6EC393}63.8 \\\midrule
\cellcolor[HTML]{FFFFFF} &
 \small{$ORI$ \sc({baseline)}}  &
  400 &
  \cellcolor[HTML]{FFFFFF}55.6 &
  \cellcolor[HTML]{FFFFFF}57.6 &
  \cellcolor[HTML]{FFFFFF}54.6 &
  \cellcolor[HTML]{FFFFFF}50.6 &
  \cellcolor[HTML]{FFFFFF}59.6 &
  \cellcolor[HTML]{FFFFFF}54.8 &
  \cellcolor[HTML]{FFFFFF}46.0 &
  \cellcolor[HTML]{FFFFFF}55.0 &
  \cellcolor[HTML]{FFFFFF}53.4 &
  \cellcolor[HTML]{FFFFFF}56.2 &
  \cellcolor[HTML]{FFFFFF}55.2 &
  \cellcolor[HTML]{FFFFFF}54.8 &
  \cellcolor[HTML]{FFFFFF}59.6 \\
\cellcolor[HTML]{FFFFFF} &
  \small{$GEN_{EN}$ } &
  3.7k &
  \cellcolor[HTML]{CAEADA}58.8 &
  \cellcolor[HTML]{AFDFC8}62.4 &
  \cellcolor[HTML]{E1F3EB}56.4 &
  \cellcolor[HTML]{E1F3EB}52.4 &
  \cellcolor[HTML]{E1F3EB}61.4 &
  \cellcolor[HTML]{C0E6D3}58.6 &
  \cellcolor[HTML]{A5DBC0}52.2 &
  \cellcolor[HTML]{F3BEB9}52.0 &
  \cellcolor[HTML]{90D2AF}63.4 &
  \cellcolor[HTML]{ABDDC5}61.2 &
  \cellcolor[HTML]{EBF7F2}56.4 &
  \cellcolor[HTML]{AFDFC8}59.6 &
  \cellcolor[HTML]{CAEADA}62.8 \\
\cellcolor[HTML]{FFFFFF} &
  \small{$GEN_{EN}$  + $ORI$}  &
  4.1k &
  \cellcolor[HTML]{B9E3CF}59.8 &
  \cellcolor[HTML]{A5DBC0}63.8 &
  \cellcolor[HTML]{A0D9BC}61.6 &
  \cellcolor[HTML]{EFF9F4}51.6 &
  \cellcolor[HTML]{CDEBDD}62.6 &
  \cellcolor[HTML]{ABDDC5}59.8 &
  \cellcolor[HTML]{AADDC5}51.2 &
  \cellcolor[HTML]{F2BCB7}51.6 &
  \cellcolor[HTML]{A0D9BC}60.4 &
  \cellcolor[HTML]{A9DDC4}61.6 &
  \cellcolor[HTML]{A3DABE}61.8 &
  \cellcolor[HTML]{90D2AF}64.8 &
  \cellcolor[HTML]{D7EFE4}62.0 \\
\cellcolor[HTML]{FFFFFF} &
  \small{$GEN_{EN}$ + $ORI$ \sc{(tlv)}}  &
  4.1k &
  \cellcolor[HTML]{ABDDC5}60.7 &
  \cellcolor[HTML]{A8DCC3}63.2 &
  \cellcolor[HTML]{A0D9BC}61.6 &
  \cellcolor[HTML]{F2FAF6}51.4 &
  \cellcolor[HTML]{AADDC5}64.8 &
  \cellcolor[HTML]{A4DABF}61.2 &
  \cellcolor[HTML]{AADDC5}51.2 &
  \cellcolor[HTML]{F9E0DE}53.6 &
  \cellcolor[HTML]{94D4B2}62.6 &
  \cellcolor[HTML]{A2D9BD}63.0 &
  \cellcolor[HTML]{CDEBDD}58.2 &
  \cellcolor[HTML]{A5DBC0}61.0 &
  \cellcolor[HTML]{A0D9BC}66.6 \\
\cellcolor[HTML]{FFFFFF} &
  \small{$GEN_{EN}$ } &
  28.6k &
  \cellcolor[HTML]{AADDC5}60.8 &
  \cellcolor[HTML]{96D5B4}66.4 &
  \cellcolor[HTML]{D4EEE1}57.2 &
  \cellcolor[HTML]{A9DDC4}56.0 &
  \cellcolor[HTML]{A2D9BD}66.4 &
  \cellcolor[HTML]{A4DABF}61.2 &
  \cellcolor[HTML]{A0D9BC}53.0 &
  \cellcolor[HTML]{FAE5E3}53.8 &
  \cellcolor[HTML]{A3DABE}60.0 &
  \cellcolor[HTML]{A9DDC4}61.6 &
  \cellcolor[HTML]{E8F6EF}56.6 &
  \cellcolor[HTML]{A3DABE}61.4 &
  \cellcolor[HTML]{ACDEC6}64.6 \\
\cellcolor[HTML]{FFFFFF} &
  \small{$GEN_{EN}$  + $ORI$}  &
  29k &
  \cellcolor[HTML]{A3DABF}62.1 &
  \cellcolor[HTML]{A0D9BC}64.6 &
  \cellcolor[HTML]{9FD8BC}61.8 &
  \cellcolor[HTML]{FFFFFF}50.6 &
  \cellcolor[HTML]{9FD8BC}66.8 &
  \cellcolor[HTML]{96D5B4}63.6 &
  \cellcolor[HTML]{DEF2E8}48.0 &
  \cellcolor[HTML]{F5FBF9}55.6 &
  \cellcolor[HTML]{82CCA4}65.8 &
  \cellcolor[HTML]{9ED8BB}63.6 &
  \cellcolor[HTML]{DEF2E8}57.2 &
  \cellcolor[HTML]{99D5B6}63.2 &
  \cellcolor[HTML]{9FD8BC}66.8 \\
\cellcolor[HTML]{FFFFFF} &
  \small{$GEN_{EN}$ + $ORI$ \sc{(tlv)}}  &
29k &
  \cellcolor[HTML]{AADDC4}60.9 &
  \cellcolor[HTML]{96D5B4}66.4 &
  \cellcolor[HTML]{9FD8BC}61.8 &
  \cellcolor[HTML]{FFFFFF}49.8 &
  \cellcolor[HTML]{A3DABE}66.2 &
  \cellcolor[HTML]{ABDDC5}59.8 &
  \cellcolor[HTML]{98D5B5}54.6 &
  \cellcolor[HTML]{F8DCD9}53.4 &
  \cellcolor[HTML]{95D4B3}62.4 &
  \cellcolor[HTML]{9DD7BA}63.8 &
  \cellcolor[HTML]{CDEBDD}58.2 &
  \cellcolor[HTML]{9BD6B8}62.8 &
  \cellcolor[HTML]{A5DBC0}65.8 \\
\cellcolor[HTML]{FFFFFF} &
  \small{$GEN_{XX}$ }&
  3.6k/lang &
  \cellcolor[HTML]{CAEADA}58.8 &
  \cellcolor[HTML]{AFDFC8}62.4 &
  \cellcolor[HTML]{D7EFE4}57.0 &
  \cellcolor[HTML]{ABDDC5}55.6 &
  \cellcolor[HTML]{E1F3EB}61.4 &
  \cellcolor[HTML]{B9E3CF}59.0 &
  \cellcolor[HTML]{92D3B1}55.6 &
  \cellcolor[HTML]{FCF2F1}54.4 &
  \cellcolor[HTML]{C6E8D8}56.8 &
  \cellcolor[HTML]{FFFFFF}-- &
  \cellcolor[HTML]{FFFFFF}-- &
  \cellcolor[HTML]{A7DCC2}60.6 &
  \cellcolor[HTML]{D7EFE4}62.0 \\
\cellcolor[HTML]{FFFFFF} &
  \small{\small{$GEN_{XX}$ } + $ORI$} &
4k &
  \cellcolor[HTML]{B7E2CE}59.9 &
  \cellcolor[HTML]{A5DBC0}63.8 &
  \cellcolor[HTML]{A6DBC1}60.6 &
  \cellcolor[HTML]{B6E2CC}55.0 &
  \cellcolor[HTML]{ACDEC6}64.6 &
  \cellcolor[HTML]{AFDFC8}59.6 &
  \cellcolor[HTML]{A3DABE}52.6 &
  \cellcolor[HTML]{FDF6F5}54.6 &
  \cellcolor[HTML]{CDEBDD}56.4 &
  \cellcolor[HTML]{FFFFFF}-- &
  \cellcolor[HTML]{FFFFFF}-- &
  \cellcolor[HTML]{AFDFC8}59.6 &
  \cellcolor[HTML]{AADDC5}64.8 \\
\cellcolor[HTML]{FFFFFF} &
  \small{$GEN_{EN}^{Trans} + ORI$} &
4k &
  \cellcolor[HTML]{A9DCC3}61.1 &
  \cellcolor[HTML]{A5DBC0}63.8 &
  \cellcolor[HTML]{A9DDC4}60.0 &
  \cellcolor[HTML]{9ED8BB}58.0 &
  \cellcolor[HTML]{A9DDC4}65.0 &
  \cellcolor[HTML]{A6DBC1}60.8 &
  \cellcolor[HTML]{FFFFFF}-- &
  \cellcolor[HTML]{FAE5E3}53.8 &
  \cellcolor[HTML]{A2D9BD}60.2 &
  \cellcolor[HTML]{90D2AF}66.2 &
  \cellcolor[HTML]{E8F6EF}56.6 &
  \cellcolor[HTML]{9CD7B9}62.6 &
  \cellcolor[HTML]{A4DABF}66.0 \\
\multirow{-11}{*}{\cellcolor[HTML]{FFFFFF}\small{XLMR-BASE}} &
  \small{$GEN_{EN}^{Trans} + ORI$} &
  29k/lang &
  \cellcolor[HTML]{A3DABE}\textbf{62.2} &
  \cellcolor[HTML]{A0D9BC}64.6 &
  \cellcolor[HTML]{98D5B5}63.2 &
  \cellcolor[HTML]{A3DABE}57.2 &
  \cellcolor[HTML]{AADDC5}64.8 &
  \cellcolor[HTML]{A4DABF}61.2 &
  \cellcolor[HTML]{FFFFFF}-- &
  \cellcolor[HTML]{FFFFFF}55.0 &
  \cellcolor[HTML]{9CD7B9}61.2 &
  \cellcolor[HTML]{CDEBDD}59.2 &
  \cellcolor[HTML]{B7E2CE}59.5 &
  \cellcolor[HTML]{93D3B1}64.2 &
  \cellcolor[HTML]{96D5B4}68.4 \\\midrule
\cellcolor[HTML]{FFFFFF} &
\small{$ORI$ \sc({baseline)}} &
  400 &
  \cellcolor[HTML]{FFFFFF}64.4 &
  \cellcolor[HTML]{FFFFFF}71.4 &
  \cellcolor[HTML]{FFFFFF}62.8 &
  \cellcolor[HTML]{FFFFFF}51.4 &
  \cellcolor[HTML]{FFFFFF}69.0 &
  \cellcolor[HTML]{FFFFFF}65.8 &
  \cellcolor[HTML]{FFFFFF}52.0 &
  \cellcolor[HTML]{FFFFFF}60.6 &
  \cellcolor[HTML]{FFFFFF}62.0 &
  \cellcolor[HTML]{FFFFFF}64.0 &
  \cellcolor[HTML]{FFFFFF}61.2 &
  \cellcolor[HTML]{FFFFFF}69.4 &
  \cellcolor[HTML]{FFFFFF}66.8 \\
\cellcolor[HTML]{FFFFFF} &
  \small{$GEN_{EN}$ } &
  3.7k &
  \cellcolor[HTML]{F2FAF6}65.2 &
  \cellcolor[HTML]{FFFFFF}71.2 &
  \cellcolor[HTML]{E1F3EB}64.6 &
  \cellcolor[HTML]{FCFEFD}51.6 &
  \cellcolor[HTML]{E1F3EB}70.8 &
  \cellcolor[HTML]{F2FAF6}66.6 &
  \cellcolor[HTML]{FBE9E7}51.0 &
  \cellcolor[HTML]{F7D8D5}58.8 &
  \cellcolor[HTML]{BCE4D1}66.0 &
  \cellcolor[HTML]{B9E3CF}68.2 &
  \cellcolor[HTML]{9CD7B9}69.0 &
  \cellcolor[HTML]{FCF2F1}68.8 &
  \cellcolor[HTML]{DEF2E8}68.8 \\
\cellcolor[HTML]{FFFFFF} &
  \small{$GEN_{EN}$  + $ORI$}  &
  4.1k &
  \cellcolor[HTML]{ABDDC5}69.5 &
  \cellcolor[HTML]{ABDDC5}76.4 &
  \cellcolor[HTML]{A0D9BC}69.8 &
  \cellcolor[HTML]{F2BDB8}48.2 &
  \cellcolor[HTML]{A0D9BC}76.0 &
  \cellcolor[HTML]{A0D9BC}72.8 &
  \cellcolor[HTML]{FAE5E3}50.8 &
  \cellcolor[HTML]{D0ECDF}63.4 &
  \cellcolor[HTML]{A7DCC2}67.8 &
  \cellcolor[HTML]{A2D9BD}70.8 &
  \cellcolor[HTML]{95D4B3}70.2 &
  \cellcolor[HTML]{BCE4D1}73.4 &
  \cellcolor[HTML]{8ACFAA}77.8 \\
\cellcolor[HTML]{FFFFFF} &
  \small{$GEN_{EN}$ + $ORI$ \sc{(tlv)}}  &
  4.1k &
  \cellcolor[HTML]{9ED8BA}71.9 &
  \cellcolor[HTML]{94D4B2}80.6 &
  \cellcolor[HTML]{96D5B4}71.6 &
  \cellcolor[HTML]{FCF2F1}50.8 &
  \cellcolor[HTML]{92D3B1}78.6 &
  \cellcolor[HTML]{88CEA8}77.2 &
  \cellcolor[HTML]{FEFAFA}51.8 &
  \cellcolor[HTML]{D7EFE4}63.0 &
  \cellcolor[HTML]{9FD8BB}69.2 &
  \cellcolor[HTML]{9FD8BB}71.2 &
  \cellcolor[HTML]{87CEA8}72.8 &
  \cellcolor[HTML]{9CD7B9}77.2 &
  \cellcolor[HTML]{85CDA6}78.8 \\
\cellcolor[HTML]{FFFFFF} &
  \small{$GEN_{EN}$ } &
  28.6k &
  \cellcolor[HTML]{9ED8BB}71.8 &
  \cellcolor[HTML]{94D4B2}80.6 &
  \cellcolor[HTML]{87CEA7}74.4 &
  \cellcolor[HTML]{FDF6F5}51.0 &
  \cellcolor[HTML]{93D3B1}78.4 &
  \cellcolor[HTML]{93D3B1}75.2 &
  \cellcolor[HTML]{FBEDEC}51.2 &
  \cellcolor[HTML]{D0ECDF}63.4 &
  \cellcolor[HTML]{9CD7B9}69.8 &
  \cellcolor[HTML]{A3DABE}70.6 &
  \cellcolor[HTML]{98D5B5}69.8 &
  \cellcolor[HTML]{A5DBC0}75.6 &
  \cellcolor[HTML]{8CD0AC}77.4 \\
\cellcolor[HTML]{FFFFFF} &
  \small{$GEN_{EN}$  + $ORI$}  &
  29k &
  \cellcolor[HTML]{9BD6B8}\textbf{72.4} &
  \cellcolor[HTML]{92D3B1}81.0 &
  \cellcolor[HTML]{8ACFAA}73.8 &
  \cellcolor[HTML]{CDEBDD}54.4 &
  \cellcolor[HTML]{89CFA9}80.2 &
  \cellcolor[HTML]{93D3B1}75.2 &
  \cellcolor[HTML]{F2BDB8}48.8 &
  \cellcolor[HTML]{F2FAF6}61.4 &
  \cellcolor[HTML]{99D5B6}70.4 &
  \cellcolor[HTML]{91D2B0}73.8 &
  \cellcolor[HTML]{94D4B2}70.4 &
  \cellcolor[HTML]{A5DBC0}75.6 &
  \cellcolor[HTML]{7FCBA1}79.8 \\
\cellcolor[HTML]{FFFFFF} &
  \small{$GEN_{EN}$ + $ORI$ \sc{(tlv)}}  &
29k &
  \cellcolor[HTML]{9BD6B8}\textbf{72.4} &
  \cellcolor[HTML]{92D3B1}81.0 &
  \cellcolor[HTML]{8ACFAA}73.8 &
  \cellcolor[HTML]{CDEBDD}54.4 &
  \cellcolor[HTML]{89CFA9}80.2 &
  \cellcolor[HTML]{93D3B1}75.2 &
  \cellcolor[HTML]{F2BDB8}48.8 &
  \cellcolor[HTML]{F9FDFB}61.0 &
  \cellcolor[HTML]{99D5B6}70.4 &
  \cellcolor[HTML]{91D2B0}73.8 &
  \cellcolor[HTML]{94D4B2}70.4 &
  \cellcolor[HTML]{A5DBC0}75.6 &
  \cellcolor[HTML]{7FCBA1}79.8 \\
\cellcolor[HTML]{FFFFFF} &
  \small{$GEN_{XX}$ }&
  3.6k/lang &
  \cellcolor[HTML]{FBE9E7}63.4 &
  \cellcolor[HTML]{FEFAFA}71.2 &
  \cellcolor[HTML]{FEFAFA}62.6 &
  \cellcolor[HTML]{D0ECDF}54.2 &
  \cellcolor[HTML]{DEF2E8}71.0 &
  \cellcolor[HTML]{FFFFFF}65.8 &
  \cellcolor[HTML]{F4C6C2}49.4 &
  \cellcolor[HTML]{F0AFA9}53.8 &
  \cellcolor[HTML]{F1B3AE}56.4 &
  \cellcolor[HTML]{FFFFFF}-- &
  \cellcolor[HTML]{FFFFFF}-- &
  \cellcolor[HTML]{F1B4AF}64.0 &
  \cellcolor[HTML]{AFDFC8}71.6 \\
\cellcolor[HTML]{FFFFFF} &
  \small{\small{$GEN_{XX}$ } + $ORI$} &
4k &
  \cellcolor[HTML]{F2FAF6}65.2 &
  \cellcolor[HTML]{ABDDC5}76.4 &
  \cellcolor[HTML]{FDF6F5}62.4 &
  \cellcolor[HTML]{C0E6D3}55.2 &
  \cellcolor[HTML]{A6DBC1}75.0 &
  \cellcolor[HTML]{F2BBB6}62.2 &
  \cellcolor[HTML]{DEF2E8}54.0 &
  \cellcolor[HTML]{F5CBC7}58.2 &
  \cellcolor[HTML]{F0B0AA}55.4 &
  \cellcolor[HTML]{FFFFFF}-- &
  \cellcolor[HTML]{FFFFFF}-- &
  \cellcolor[HTML]{F2BDB8}66.2 &
  \cellcolor[HTML]{93D3B1}76.2 \\
\cellcolor[HTML]{FFFFFF} &
  \small{$GEN_{EN}^{Trans} + ORI$} &
4k &
  \cellcolor[HTML]{D4EEE1}67.0 &
  \cellcolor[HTML]{ABDDC5}76.4 &
  \cellcolor[HTML]{F3C2BD}60.0 &
  \cellcolor[HTML]{9AD6B7}59.6 &
  \cellcolor[HTML]{F3C2BD}66.2 &
  \cellcolor[HTML]{F2FAF6}66.6 &
  \cellcolor[HTML]{FFFFFF}-- &
  \cellcolor[HTML]{F8DCD9}59.0 &
  \cellcolor[HTML]{D0ECDF}64.8 &
  \cellcolor[HTML]{9FD8BB}71.2 &
  \cellcolor[HTML]{BCE4D1}65.2 &
  \cellcolor[HTML]{A9DDC4}74.8 &
  \cellcolor[HTML]{96D5B4}75.6 \\
\multirow{-11}{*}{\cellcolor[HTML]{FFFFFF}\small{XLMR-LARGE}} &
  \small{$GEN_{EN}^{Trans} + ORI$} &
  29k/lang &
  \cellcolor[HTML]{A0D9BC}71.5 &
  \cellcolor[HTML]{92D3B1}81.0 &
  \cellcolor[HTML]{95D4B3}71.8 &
  \cellcolor[HTML]{A7DCC2}57.2 &
  \cellcolor[HTML]{8BD0AB}79.8 &
  \cellcolor[HTML]{98D5B5}74.4 &
  \cellcolor[HTML]{FFFFFF}-- &
  \cellcolor[HTML]{F0B3AD}54.8 &
  \cellcolor[HTML]{93D3B1}71.4 &
  \cellcolor[HTML]{98D5B5}72.6 &
  \cellcolor[HTML]{96D5B4}70.0 &
  \cellcolor[HTML]{9CD7B9}77.2 &
  \cellcolor[HTML]{96D5B4}75.6 \\
\bottomrule
\end{tabular}
}
\caption{Full results on XCOPA (with ChatGPT-generated data).
\textsc{+tlv} corresponds to including the original validation set in all \textsc{\textbf{t}}arget \textsc{\textbf{l}}anguages in the \textsc{\textbf{v}}alidation set.
Rows are sorted by the number of instances used in training.
\textsc{avg} shows average results for languages that are available in all settings (excl. QU, TH, TR).
}
 \label{tab:xcopa_full}
\end{table*}

\begin{table*}[t]
\centering
\scalebox{0.8}{
\addtolength{\tabcolsep}{-1.5pt}

\begin{tabular}{ll|ccccccccccccc}
\toprule
\textbf{Model} &
  \textbf{Training Data} &
  \textbf{AVG} &
  \textbf{EN} &
  \textbf{ET} &
  \textbf{HT} &
  \textbf{ID} &
  \textbf{IT} &
  \textbf{QU} &
  \textbf{SW} &
  \textbf{TA} &
  \textbf{TH} &
  \textbf{TR} &
  \textbf{VI} &
  \textbf{ZH} \\
  \midrule
\multirow{7}{*}{mBERT} &
  \small{$ORI$} &
47.2
& 53.8 & 44.2 & 48.6 & 47.2 & 46.2 & 50.6 & 45.4 & 48.4 & 49.8 & 49.8 & 43.6 & 47.4 \\
&
  \small{$GEN_{EN}$} &
  58.2 &
  69.2 &
  59.2 &
  54.0 &
  60.6 &
  59.2 &
  50.8 &
  48.2 &
  55.0 &
  48.2 &
  53.8 &
  57.6 &
  61.0 \\
 &
  \small{$GEN_{EN}+ORI$} &
  59.3 &
  72.6 &
  58.8 &
  53.0 &
  62.0 &
  61.0 &
  53.0 &
  50.0 &
  54.0 &
  48.2 &
  52.0 &
  57.6 &
  64.6 \\
 &
  \small{$GEN_{XX}$} &
  60.2 &
  69.2 &
  59.4 &
  56.2 &
  60.2 &
  63.8 &
  54.4 &
  55.2 &
   54.0 &
  -- &
  -- &
  61.2 &
  62.2 \\
 &
  \small{$GEN_{XX}+ORI$} &
  61.8 &
  72.6 &
  61.2 &
  58.2 &
  62.2 &
  66.4 &
  54.4 &
  57.4 &
  53.4 &
  -- &
  -- &
  63.0 &
  61.8 \\
 &
  \small{$GEN_{EN}^{Trans}$} &
  61.4 &
  69.2 &
  59.2 &
  56.8 &
  65.4 &
  65.2 &
  -- &
  53.4 &
  56.8 &
  52.6 &
  59.6 &
  61.8 &
  65.0 \\
 &
  \small{$GEN_{EN}^{Trans}+ORI$} &
  62.6 &
  72.6 &
  58.6 &
  55.2 &
  65.6 &
  65.4 &
  -- &
  53.8 &
  62.6 &
  53.2 &
  58.8 &
  64.6 &
  65.4 \\
  \midrule
\multirow{7}{*}{XLMR-Base} &
  \small{$ORI$} &
  55.6 &
  57.6 &
  54.6 &
  50.6 &
  59.6 &
  54.8 &
  46.0 &
  55.0 &
  53.4 &
  56.2 &
  55.2 &
  54.8 &
  59.6 \\
 &
  \small{$GEN_{EN}$} &
  63.6 &
  67.0 &
  62.4 &
  52.0 &
  68.6 &
  62.6 &
  51.8 &
  58.6 &
  65.4 &
  64.8 &
  63.2 &
  66.6 &
  69.6 \\
 &
  \small{$GEN_{EN}+ORI$} &
  63.6 &
  69.6 &
  63.8 &
  51.2 &
  67.2 &
  62.4 &
  52.6 &
  58.4 &
  63.8 &
  66.0 &
  64.2 &
  66.8 &
  69.4 \\
 &
  \small{$GEN_{XX}$} &
  63.2 &
  67.0 &
  60.8 &
  56.4 &
  68.6 &
  62.4 &
  57.4 &
  58.2 &
  60.2 &
  -- &
  -- &
  64.6 &
  70.4 \\
 &
  \small{$GEN_{XX}+ORI$} &
  64.0 &
  69.6 &
  62.2 &
  56.2 &
  68.6 &
  63.8 &
  56.8 &
  57.8 &
  61.2 &
  -- &
  -- &
  66.8 &
  70.0 \\
 &
  \small{$GEN_{EN}^{Trans}$} &
  62.5 &
  67.0 &
  60.0 &
  55.6 &
  66.0 &
  62.4 &
  -- &
  58.0 &
  60.4 &
  64.4 &
  64.6 &
  64.0 &
  68.8 \\
 &
  \small{$GEN_{EN}^{Trans}+ORI$} &
  63.9 &
  69.6 &
  61.6 &
  56.6 &
  68.4 &
  65.2 &
  -- &
  58.2 &
  60.2 &
  68.0 &
  62.6 &
  66.0 &
  69.6 \\
  \midrule
\multirow{7}{*}{XLMR-Large} &
  \small{$ORI$} &
  64.4 &
  71.4 &
  62.8 &
  51.4 &
  69.0 &
  65.8 &
  52.0 &
  60.6 &
  62.0 &
  64.0 &
  61.2 &
  69.4 &
  66.8 \\
 &
  \small{$GEN_{EN}$} &
  73.6 &
  83.2 &
  71.2 &
  52.0 &
  81.2 &
  78.2 &
  51.0 &
  62.2 &
  76.6 &
  77.4 &
  75.0 &
  78.4 &
  79.0 \\
 &
  \small{$GEN_{EN}+ORI$} &
  73.7 &
  84.6 &
  70.4 &
  50.0 &
  80.8 &
  80.2 &
  51.8 &
  65.8 &
  72.8 &
  76.0 &
  74.8 &
  78.4 &
  80.4 \\
 &
  \small{$GEN_{XX}$} &
  72.8 &
  83.2 &
  75.2 &
  55.2 &
  78.4 &
  76.0 &
  52.4 &
  63.0 &
  68.2 &
  -- &
  -- &
  77.8 &
  78.6 \\
 &
  \small{$GEN_{XX}+ORI$} &
  74.6 &
  84.6 &
  77.0 &
  56.0 &
  82.2 &
  77.0 &
  56.0 &
  65.0 &
  73.8 &
  -- &
  -- &
  76.2 &
  80.0 \\
 &
  \small{$GEN_{EN}^{Trans}$} &
  71.0 &
  83.2 &
  72.4 &
  55.6 &
  79.4 &
  78.2 &
  -- &
  60.6 &
  67.8 &
  77.8 &
  72.6 &
  64.0 &
  77.4 \\
 &
  \small{$GEN_{EN}^{Trans}+ORI$} &
  74.1 &
  84.6 &
  74.2 &
  57.2 &
  82.0 &
  77.4 &
  -- &
  62.2 &
  75.0 &
  75.2 &
  72.8 &
  74.4 &
  79.6 \\
  \bottomrule
\end{tabular}
}
\caption{Accuracy on XCOPA. {\footnotesize{$GEN_{EN}$}} and {\footnotesize{$GEN_{XX}$}} represents 3.7K and 3.6K data in English and target languages generated by GPT-4.
\textsc{avg} shows average results for languages that are available in all settings (excl. QU, TH, TR).}
 \label{tab:gpt4-copa}
\end{table*}

\end{document}